\documentclass[sigconf]{acmart}

\AtBeginDocument{%
  \providecommand\BibTeX{{%
    Bib\TeX}}}

\setcopyright{acmlicensed}
\copyrightyear{2018}
\acmYear{2018}
\acmDOI{XXXXXXX.XXXXXXX}
\acmConference[Conference acronym 'XX]{Make sure to enter the correct
  conference title from your rights confirmation email}{June 03--05,
  2018}{Woodstock, NY}
\acmISBN{978-1-4503-XXXX-X/2018/06}

\usepackage{pgfplots}
\pgfplotsset{compat=1.17}
\usepgfplotslibrary{groupplots}

\usepackage{amsmath,amsfonts}
\usepackage{bbm}
\usepackage{amsthm}
\usepackage{algorithmic}
\usepackage{graphicx}
\usepackage{textcomp}
\usepackage{xcolor}
\usepackage[table,xcdraw]{xcolor}
\usepackage{booktabs}
\usepackage{multirow} 
\usepackage{pifont}
\def\BibTeX{{\rm B\kern-.05em{\sc i\kern-.025em b}\kern-.08em
    T\kern-.1667em\lower.7ex\hbox{E}\kern-.125emX}}

\definecolor{darkgreen}{rgb}{0.0, 0.5, 0.0}
\definecolor{darkblue}{RGB}{31, 119, 180}

\definecolor{PaleGreen}{HTML}{E8FAE8}
\definecolor{PaleRed}{HTML}{FFEBEB}     
\definecolor{PaleGrey}{HTML}{F5F7FA}
\definecolor{PosCellGreen}{HTML}{C6F7C6}

\definecolor{DIblue1}{HTML}{5393D5} 
\definecolor{DIblue2}{HTML}{7EA9E1}
\definecolor{DIblue3}{HTML}{A6C8ED}
\definecolor{DIblue4}{HTML}{D0E3F7}
\definecolor{DIblue5}{HTML}{EAF2FB} 
\definecolor{Green0}{HTML}{F7FCF9} 
\definecolor{Green1}{HTML}{E9F9ED} 
\definecolor{Green1.5}{HTML}{CDEFD7} 
\definecolor{Green2}{HTML}{B7E4C7} 
\definecolor{Green2.5}{HTML}{86C89C}
\definecolor{Green3}{HTML}{51B37F} 

\definecolor{customcyan}{RGB}{0, 158, 115} 
\definecolor{tealblue}{RGB}{0, 114, 178}
\definecolor{darkorange}{RGB}{213, 94, 0} 

\newcommand{\std}[1]{\footnotesize{\color{gray}$\pm$#1}}

\theoremstyle{remark}

\begin{document}

\title{Training Diverse Graph Experts for Ensembles: A Systematic Empirical Study}

\author{Gangda Deng}
\authornote{Equal contribution.}
\affiliation{%
  \institution{University of Southern California}
  \city{Los Angeles}
  \country{USA}
}
\email{gangdade@usc.edu}

\author{Yuxin Yang}
\authornotemark[1]
\affiliation{%
  \institution{University of Southern California}
  \city{Los Angeles}
  \country{USA}
}
\email{yyang393@usc.edu}

\author{Ömer Faruk Akgül}
\authornotemark[1]
\affiliation{%
  \institution{University of Southern California}
  \city{Los Angeles}
  \country{USA}
}
\email{akgul@usc.edu}

\author{Hanqing Zeng}
\affiliation{%
  \institution{Meta AI}
  \city{Menlo Park}
  \country{USA}
}
\email{zengh@meta.com}

\author{Yinglong Xia}
\affiliation{%
  \institution{Meta AI}
  \city{Menlo Park}
  \country{USA}
}
\email{yxia@meta.com}

\author{Rajgopal Kannan}
\affiliation{%
  \institution{DEVCOM ARL Army Research Office}
  \city{Los Angeles}
  \country{USA}
}
\email{rajgopal.kannan.civ@army.mil}

\author{Viktor Prasanna}
\affiliation{%
  \institution{University of Southern California}
  \city{Los Angeles}
  \country{USA}
}
\email{prasanna@usc.edu}

\begin{abstract}

Graph Neural Networks (GNNs) have become essential tools for learning on relational data, yet the performance of a single GNN is often limited by the heterogeneity present in real-world graphs. Recent advances in Mixture-of-Experts (MoE) frameworks demonstrate that assembling multiple, explicitly diverse GNNs with distinct generalization patterns can significantly improve performance. In this work, we present the first systematic empirical study of expert-level diversification techniques for GNN ensembles. Evaluating 20 diversification strategies---including random re-initialization, hyperparameter tuning, architectural variation, directionality modeling, and training data partitioning---across 14 node classification benchmarks, we construct and analyze over 200 ensemble variants.
Our comprehensive evaluation examines each technique in terms of expert diversity, complementarity, and ensemble performance. We also uncovers mechanistic insights into training maximally diverse experts. These findings provide actionable guidance for expert training and the design of effective MoE frameworks on graph data. Our code is available at \url{https://github.com/Hydrapse/bench-gnn-diversification}.




\end{abstract}



\maketitle

\section{Introduction}
Graph Neural Networks (GNNs) have emerged as the standard approach for learning on relational data.
Over the years, researchers have developed numerous techniques to enhance GNN performance in node classification from both architectural and data perspectives. These techniques address significant challenges posed by the high heterogeneity of data samples in complex, irregular graphs. Such challenges include over-smoothing when leveraging higher-order neighborhoods~\cite{gcnii, ndls}, handling non-homophilous networks~\cite{acmgcn, gprgnn}, and exploiting edge directionality~\cite{sun2024breaking-357,dirgnn}.

Despite the success of carefully tuned GNNs in fitting complex graph data, existing approaches often face severe generalization bottlenecks due to limited graph size and the imbalanced distribution of node samples~\cite{subgroup, demystifying, moscat}. These constraints can undermine the advantages of increasingly powerful models, leading to diminishing returns with further parameterization. For instance, replacing conventional message-passing architectures~\cite{message-passing} with more expressive Transformer architectures~\cite{vaswani2017attention-20d} yields no performance improvements across numerous graph benchmarks~\cite{tuned, luo2025can}.

Recent advances have shifted toward Mixture-of-Experts (MoE) frameworks as a promising way to overcome these limitations~\cite{graphmetro,ma2024mixture-d0d,mowst,chen2025mixturedecoupledmessagepassing,moscat}. By using a learnable gating model to combine GNN experts, each exhibiting distinct generalization behaviors, MoE methods achieve superior aggregate performance. Notably, the “train-then-merge” design~\cite{li2022branch}, where each expert is trained independently before gating, has proven especially effective on graphs~\cite{ma2024mixture-d0d, moscat}. However, the success of these ensembles fundamentally relies on experts to specialize in different structural patterns and to make non-overlapping errors~\cite{moscat,graphmoe}. This raises a central and widely relevant question for model ensemble:
\textit{How can we systematically train diverse experts that make complementary errors?}

In this work, we present the first systematic empirical study of expert diversification techniques for GNN ensembles. We organize and benchmark 20 diversification strategies spanning five key dimensions: random reinitialization, hyperparameter tuning, architectural variation, directional modeling, and training data selection. Our large-scale evaluation covers over 200 ensemble variants across 14 widely used node classification datasets, encompassing diverse graph types, sizes, homophily levels, and edge directionality. Our main findings are as follows:

\textbf{\ding{172} Relationship among Diversity, Complementarity, and Ensemble Performance.} 
We observe that greater diversity among experts typically leads to enhanced complementarity and improved ensemble performance. 
When an ensemble fails to meet expectations despite high diversity, the problem may stem from a suboptimal gating model, not from the diversification techniques. 





\textbf{\ding{173} Best Practice for Diversifying Experts.}
Model pairs that diverge more in training strategies produce greater diversity (reinitializations $<$ directional modeling $<$ hyperparameters $<$  architectures $<$ training data selection):
(1) \textit{Hyperparameters and Architectures}: 
Dropout tuning and shallow versus deep architectures produce strong diversification; Laplacian filters and GraphSAGE-style skip connections further enhance diversity.
(2) \textit{Directional modeling}: Complementary gains from mixing directed and undirected GNNs are limited.
(3) \textit{Training set selection}: Training data partitioning yields the most distinctly specialized experts, especially when using intra-class graph metrics (i.e., heuristics calculated on filtered graphs where only intra-class edges are preserved).

\textbf{\ding{174} Mechanistic Insights behind Expert Diversification.}
(1) We introduce the Direction Informativeness (DI) metric to identify when edge directionality benefits expert diversity. Both DI and empirical results show that, directed GNNs significantly diverge from undirected variants on only 2 of 7 digraph datasets---far fewer than previously reported. 
(2) We introduce the use of intra-class graph metrics to guide training data partitioning. Rather than full-graph metrics, using intra-class graph metrics leads to more pronounced label distribution shifts and greater disparities in expert generalization. 
(3) Experts pretrained on the full training set and subsequently fine-tuned on node subgroups achieve higher overall ensemble performance but exhibit lower complementarity compared to experts trained exclusively on their respective subgroups.



\textbf{\ding{175} Downstream Task Performance.} By restricting to a maximum of three experts and employing a single diversification method, our ensembles achieve a 0.7\%$\sim$3.9\% improvement in node classification accuracy over the strongest individual model, demonstrating that principled expert diversification can consistently boost GNN performance across a broad range of settings.

\section{Problem Definition and Setup}

\subsection{Problem Definition}
In this paper, we focus on training highly divergent experts to enhance MoE and ensemble performance on node classification tasks.
We adopt the two-stage, train-then-merge MoE setup proposed in Moscat~\cite{moscat}.
Specifically, $K$ GNN experts $\{\mathcal{M}^k\}_{k=1}^{K}$ are first trained independently on training nodes.
Let $\mathbf{Z}^{(k)}$ denote the logits for expert $\mathcal{M}^{k}$, a gating model $\phi$ is then trained separately on a holdout (validation) set to aggregate the expert logits and produce the final predictions: $\widehat{\mathbf{Y}} = \phi\left(\mathbf{Z}^{(1)}, ..., \mathbf{Z}^{(K)}\right)$. 
Our goal is to systematically evaluate a comprehensive set of diversification techniques for selecting and training experts $\{\mathcal{M}^k\}_{k=1}^{K}$.
Related work on model ensembles and graph-based MoEs is provided in Appendix~\ref{sec:relate}.




\subsection{Motivation}
\textit{Why adopt the independent train-then-merge setup instead of expert-gate joint training for MoE?}
While expert-gate joint training is the prevailing approach for LLM-based MoEs and has recently gained increasing interest in graph learning~\cite{graphmoe, mowst, graphmetro}, recent study~\cite{moscat,ma2024mixture-d0d} inspired by advancements in LLM MoEs~\cite{li2022branch,sukhbaatar2024branch} demonstrates that the train-then-merge strategy offers unique advantages for GNN MoEs. Specifically, this approach better captures the diverse generalization behaviors of individual experts and avoids the potentially harmful regularizations introduced by joint training. Beyond improved performance, the train-then-merge paradigm provides a cleaner and more interpretable framework by decoupling expert training from gating model training. This separation allows us to systematically investigate how to promote expert diversity and directly analyze their generalization behaviors.

\textit{Why is expert diversification important?}
Prior work~\cite{moscat, no-one, oshana2022efficient-cc3} has established that combining diverse experts is essential for achieving significant accuracy improvements: the less correlated the errors among experts, the stronger the ensemble or mixture performance. 
Beyond hyperparameter and architectural techniques to encourage expert diversification, we further explore the impact of training data selection on expert diversification, shedding light on how structural properties of training data influence expert generalization. Additionally, a deeper understanding of expert diversification allows us to identify model designs that excel on specific data subsets, even if their overall performance appears suboptimal and is often overshadowed in evaluations on the whole graph.

\begin{figure*}[t!]
\vspace{-10pt}
  \centering
  \includegraphics[width=0.85\textwidth]{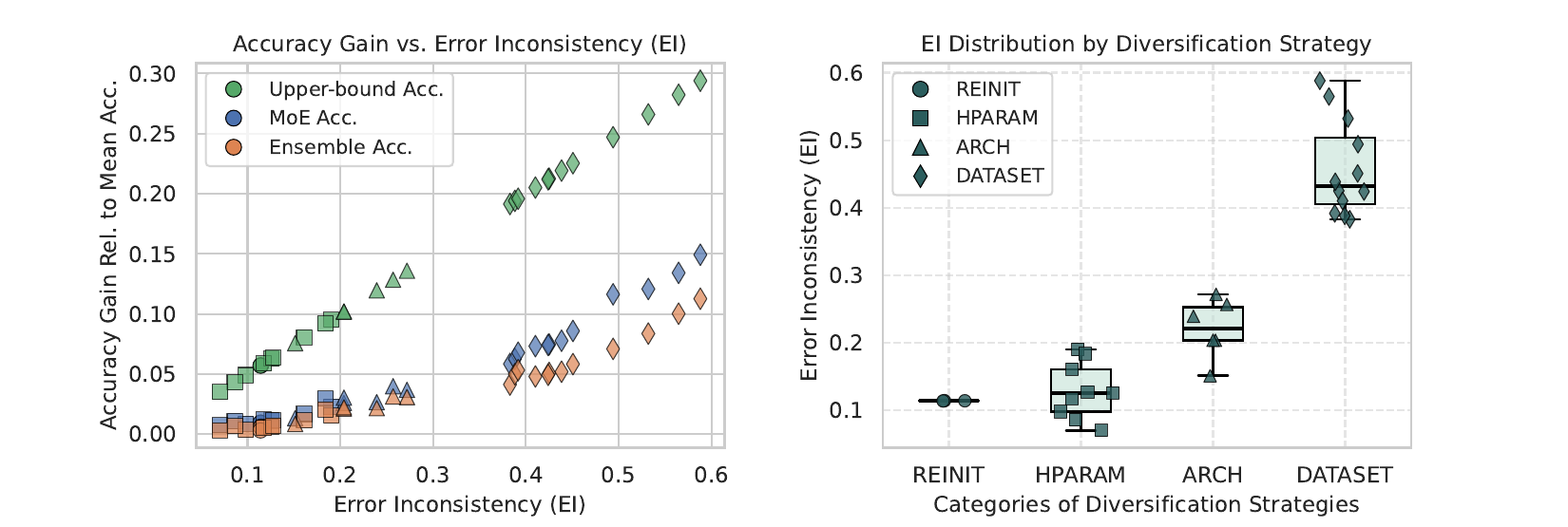} 
  \vspace{-10pt}
  \caption{%
    \textbf{Left:} The relationship between accuracy gain (relative to the mean expert accuracy) and error inconsistency (EI) in the Amazon-Ratings dataset. Each datapoint corresponds to a pair of GNNs diversified using a specific strategy. 
    We report results from three ensemble strategies: \textcolor{darkgreen!90}{\emph{Upper Bound}}, \textcolor{darkblue}{\emph{Moscat Gating Ensemble} (MoE)}, and \textcolor{orange}{\emph{Global-Ensemble} (Ensemble)}. 
    \textbf{Right:} Boxplots of EI across the four expert diversification strategy categories. 
  }
  \label{fig:upperbound_inconsistency}
  \vspace{-10pt}
\end{figure*}

\subsection{Methodology} \label{sec:methodology}

\noindent\textbf{Datasets.}  
We benchmark 14 real-world graphs that span various homophily levels and scales. Tiny graphs (with $<\!1$k nodes) and synthetic-edge graphs are excluded for robustness~\cite{critical}. See Appendix~\ref{app:datasets} for the full dataset list.

\noindent\textbf{Expert Architectures.} Our study focuses on message-passing GNN architectures, where we systematically evaluate the effects of different modular components by varying one at a time within a unified framework (see Appendix~\ref{app:expert_arch} for details).








\noindent\textbf{Evaluation.}
We evaluate the effectiveness of expert diversification techniques for node classification from three perspectives: \textit{Diversity}, \textit{Complementarity}, and \textit{Ensemble/MoE Accuracy}. Each aspect captures a distinct property necessary for assessing the true benefits of expert diversification.

\ding{172} \textit{Diversity}:
Diversity among experts is a key factor driving performance gains in MoE frameworks. Let $\mathcal{D} = \{(x_n, y_n)\}_{n=1}^{N}$ denote the evaluation set, and $\{\mathcal{M}^k\}_{k=1}^{K}$ the set of $K$ trained experts. The correctness indicator for expert $k$ on sample $n$ is defined as
$c_{k,n} = \mathbbm{1}\{\mathcal{M}^k(x_n) = y_n\} \in \{0, 1\}$,
where $\mathbbm{1}\{\cdot\}$ is the indicator function.
To quantify how differently the experts behave, we use \textbf{Error Inconsistency (EI)}~\cite{no-one}, which, for a pair of experts $(i, j)$, measures the probability that exactly one of them predicts correctly:
\begin{equation}
\mathrm{EI}_{i, j}
= \frac{1}{N} \sum_{n=1}^{N} |c_{i, n} - c_{j, n}|
= \Pr(c_{i} \neq c_{j}).
\end{equation}
For $K \geq 2$, we generalize EI as the proportion of nodes not classified consistently---whether correctly or incorrectly---by all experts:
\begin{equation}
\label{eq:ei-set}
\mathrm{EI}
= \frac{1}{N} \sum_{n=1}^{N}
\mathbbm{1}\left\{
\min_{k} c_{k,n} = 0
\;\land\;
\max_{k} c_{k,n} = 1
\right\},
\end{equation}
where the indicator is 1 if there is at least one correct and one incorrect prediction among the experts for sample $n$.

\ding{173} \textit{Complementarity}:
Beyond diversity, we seek experts whose errors are non-overlapping, enabling complementary strengths. We first define the \textbf{Upper-bound Accuracy} as the proportion of nodes correctly predicted by at least one expert:
\begin{equation}
\mathrm{Acc}_{\text{oracle}}
= \frac{1}{N} \sum_{n=1}^{N} \mathbbm{1}\left\{\max_{k} c_{k,n} = 1\right\}.
\end{equation}
We then define \textbf{Complementary Gain (CG)} as the improvement of this upper bound over the best single expert’s accuracy. Letting $\mathrm{Acc}_k$ denote the accuracy of expert $k$,
\begin{equation}
\mathrm{CG} = \mathrm{Acc}_{\text{oracle}} - \max_{k} \mathrm{Acc}_k.
\end{equation}
These metrics reflect the performance of oracle ensembling, sidestepping the limitations of potentially suboptimal practical ensembles.

\ding{174} \textit{Ensemble and MoE Accuracy}:
We further assess the effectiveness of expert diversification using two practical ensemble approaches. (1) \textbf{Global-Ensemble} (denoted as \textbf{Ensemble} in the following) applies a \textit{global} scalar weight to each expert for all nodes, with weights tuned on the validation set using the TPESampler from Optuna~\cite{optuna_2019}. (2) \textbf{Moscat Gating Ensemble} (denoted as \textbf{MoE}) employs a gating model that assigns \textit{node-adaptive} weights to each expert, dynamically integrating their predictions for each node based on validation performance. We evaluate the practical ensemble effectiveness of expert diversification using these two methods.

\section{Empirical Findings and Analysis}
In this section, we present results and analysis of expert diversification. We begin by giving an overview of five major categories of diversification strategies (Section~\ref{sec:overview_div}). We then conduct controlled studies on hyperparameters and architectural variations (Section ~\ref{sec:arch-search}), followed by an in-depth analysis of directional modeling (Section~\ref{sec:directionality}). We further investigate training data selection strategies (Section~\ref{sec:train_data_sel}) and propose intra-class graph metrics as a principled approach to domain discovery (Section~\ref{sec:intra-class}). Finally, we evaluate these strategies on the downstream task (Section~\ref{sec:downstream}).

\subsection{Overview of Expert Diversification Strategies} \label{sec:overview_div}

To understand how various diversification strategies impact the diversity and complementarity of experts, we organize all the strategies into the following categories:

\begin{itemize}
\item \textbf{Reinitialization (REINIT)}: Changing random seeds during training.
\item \textbf{Directional Modeling (DIRECT)}: Combining GNNs with directed and undirected modeling of the graph. 
\item \textbf{Hyperparameter tuning (HPARAM)}: Modifying dropout rates, hidden dimensions, or training epochs.
\item \textbf{Architecture variation (ARCH)}: Changing GNN filters (e.g., $A_{sym}$, $L_{sym}$) or depth (e.g., shallow vs. deep models).
\item \textbf{Training data selection (DATASET)}: Altering training subsets by clustering or heuristic-based partitioning. 
\end{itemize}

Figure~\ref{fig:upperbound_inconsistency} summarizes two key observations. On the \textit{left}, we show that ensemble accuracy gains are positively correlated with error inconsistency across three ensemble methods: Upper Bound, MoE, and Ensemble. Notably, ensembles with higher EI offer greater potential for improvement.
On the \textit{right}, a boxplot analysis reveals how different diversification techniques induce varying levels of EI. Among the four methodology categories, dataset diversification (e.g., via training nodes clustering or subsampling) yields the highest EI, followed by architectural and hyperparameter variations. 
We exclude directional modeling (DIRECT) in Figure~\ref{fig:upperbound_inconsistency} and separately discuss it in section ~\ref{sec:directionality} due to its limited effectiveness.

Our results highlight the critical role of error inconsistency in driving ensemble effectiveness: techniques that increase disagreement among experts (i.e., higher EI) also unlock greater performance gains when their predictions are aggregated.


\begin{figure}[t]
\hspace{-10pt}
  \centering
  \includegraphics[width=0.48\textwidth]{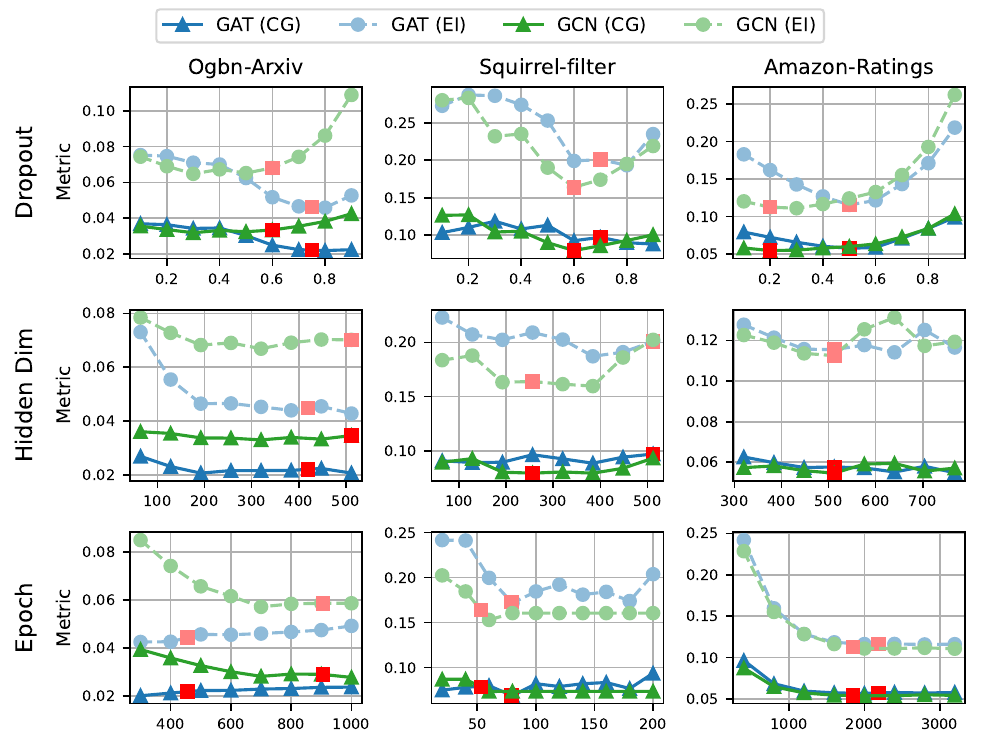}
  \caption{%
    Pairwise Error Inconsistency (EI) and Complementary Gain (CG) trend with a single modified hyperparameter (averaged over 5 random seeds). The datapoints are computed between a pair of experts: a tuned \emph{best-performing expert} as baseline and a \emph{single-hyperparameter variant} on the baseline.
  }
  \label{fig:hparam-inconsistency}
  \vspace{-15pt}
\end{figure}

\begin{figure}[t]
\hspace{-14pt}
  \centering
  \includegraphics[width=0.50\textwidth]{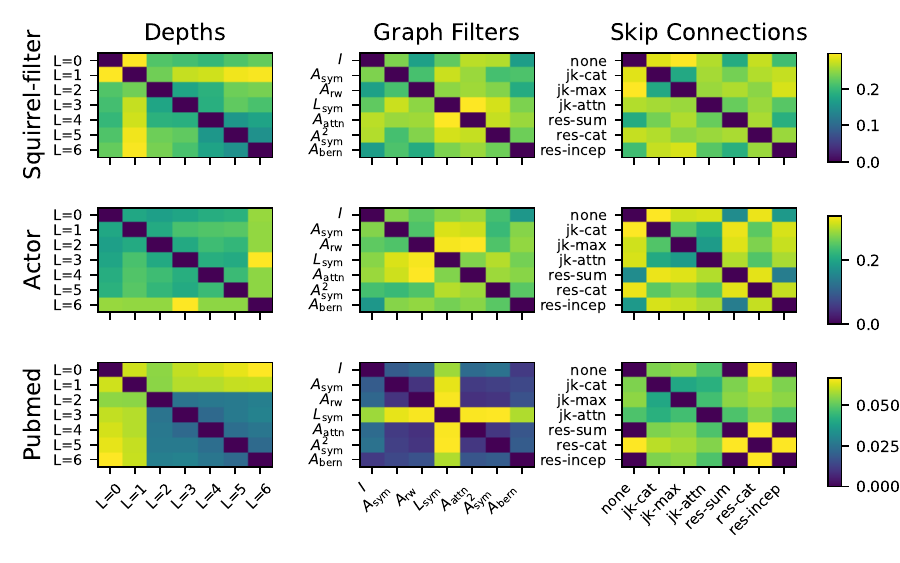}
  \vspace{-20pt}
  \caption{Pairwise error inconsistency across architectural design dimensions. Each heatmap displays the average error inconsistency (computed over 10 runs) between expert pairs that differ on a single architecture design.
   Each dataset in the same row shares a color scale.}
  \label{fig:arch-inconsistency}
\vspace{-10pt}
\end{figure}

\subsection{Architectural and Hyperparameter Search for Divergent Experts}\label{sec:arch-search}



To systematically evaluate expert diversification, we conduct a controlled search over general hyperparameters and architectural components. We first tune a base GCN or GAT following \cite{tuned}, then perturb one hyperparameter (dropout, hidden dimension, or training epochs) and compare the perturbed model to the base using error inconsistency (EI) and complementary gain (CG). 
Figure \ref{fig:hparam-inconsistency} explores the effect of hyperparameter variation and reveals a consistent U-shape: EI and CG are lowest at the tuned setting (red square) and climb as the change grows (e.g., on Amazon-Ratings, shifting dropout from 0.5 to 0.25 / 0.75 nearly doubles EI), showing that even simple hyperparameter tweaks can create substantial expert diversity; complete results are shown in Appendix \ref{app:param-ei}.
Figure~\ref{fig:arch-inconsistency} explores three architectural axes (depth, graph filters, and skip connections) using GCN as the base model. These dimensions directly influence the receptive field, information aggregation, and feature propagation mechanisms, covering a broad and representative design space in message-passing GNNs.
Our main conclusions are: The shallowest GNN ($L{=}0$) produces the largest EI; non-smoothing filters (identity matrix or high-pass Laplacian matrix) drive high EI on homophilic datasets; and skip-connection variants yield smaller, dataset-specific shifts. Complete analysis is shown in Appendix \ref{app:arch-ei}.

These results demonstrate that both hyperparameter and architectural search are effective ways to construct diverse expert sets. It also suggests that configurations yielding higher EI should be prioritized when building ensembles for potential accuracy gain.

\subsection{How can Directional Modeling Help Expert Diversification?} \label{sec:directionality}


Graph datasets can be categorized based on whether their edges are naturally directed or undirected. Classic examples of directed graphs (digraphs) include citation networks and wiki (web page) networks. In contrast, undirected graphs are typical in co-occurrence networks, such as co-author and co-purchase networks. Social networks often represent a unique case, encompassing both directed and undirected structures.

Most GNN methods assume the input graphs is undirected. When presented with naturally directed graphs, a common workaround is to ignore the directionality by symmetrizing the adjacency matrix. However, this can obscure crucial structural information. Recent study has sought to overcome this limitation by explicitly modeling directional patterns (DPs) in graphs.
Given a digraph $A_{\text{dir}}$, directional patterns correspond to $k$-order directed paths for each node, which can be captured using digraph filters. Existing directed GNNs employ filters of varying orders, typically applying different weight matrices to node features processed by these filters. For instance, DirGCN~\cite{dirgnn} uses first-order filters ($A_{\text{dir}}$ and $A_{\text{dir}}^T$), HoloNet~\cite{koke2023holonets-3f3} extends this to second-order (e.g., $A_{\text{dir}}A_{\text{dir}}$, $A_{\text{dir}}^TA_{\text{dir}}^T$, etc.), and ADPA~\cite{sun2024breaking-357} further generalizes to higher-order patterns.

So when do GNNs with directional modeling diverge from undirected GNNs in node classification tasks on real-world digraphs?
Pioneering work~\cite{dirgnn} has shown that directed GNNs excel in heterophilous digraphs, while undirected GNNs are better suited for homophilous ones. This suggests the possibility of using homophily metrics to decide when to model directionality. However, conventional homophily metrics disregard edge direction, which can be suboptimal. To address this,~\cite{sun2024breaking-357} introduced AMUD, a statistical metric that measures the Pearson correlation ($\mathrm{Corr}$) between directed topology and node labels. Specifically, first-order AMUD measures the normalized difference between "$\mathrm{Corr}($in-degree, label$)$" and "$\mathrm{Corr}($out-degree, label$)$", with larger difference indicating greater informativeness of directionality for label prediction.

\begin{table*}[!ht]
    \vspace{-10pt}
    \centering
    \setlength{\tabcolsep}{3.5pt}
    \caption{
    Comparison of domain assignment strategies with 2 domains and tuned GCN experts across datasets. $\Delta$Acc (\%) represents the weighted average accuracy gain of domain-specific experts over the full-data expert (defined in \eqref{eq:acc_gain}), while EI (\%) measures error inconsistency. Higher values indicate better performance for both metrics.    
    }
    \resizebox{0.85\textwidth}{!}{
    \begin{tabular}{llcc cc cc cc cc}
    \toprule
    \multirow{2}{*}{\textbf{Type}} & \multirow{2}{*}{\textbf{Method}} 
    & \multicolumn{2}{c}{\textbf{Arxiv}} 
    & \multicolumn{2}{c}{\textbf{Squirrel-filter}} 
    & \multicolumn{2}{c}{\textbf{Amazon-Ratings}} 
    & \multicolumn{2}{c}{\textbf{Penn94}}
    & \multicolumn{2}{c}{\textbf{Flickr}} \\
    \cmidrule(lr){3-4} \cmidrule(lr){5-6} \cmidrule(lr){7-8} \cmidrule(lr){9-10} \cmidrule(lr){11-12}
     & & $\boldsymbol{\Delta}$\textbf{Acc} (\%) & \textbf{EI} (\%) 
     & $\boldsymbol{\Delta}$\textbf{Acc} (\%) & \textbf{EI} (\%) 
     & $\boldsymbol{\Delta}$\textbf{Acc} (\%) & \textbf{EI} (\%) 
     & $\boldsymbol{\Delta}$\textbf{Acc} (\%) & \textbf{EI} (\%)
     & $\boldsymbol{\Delta}$\textbf{Acc} (\%) & \textbf{EI} (\%) \\
    \midrule
    \multicolumn{2}{c}{Random Split}        & -1.00 & 11.13 & -0.95 & 21.38 & -10.91 & 38.28 & -4.32 & 22.14 & -2.22 & 10.87 \\
    \midrule
    \multirow{3}{*}{Topology-Only}
        & \multicolumn{1}{|l}{Degree}        & -0.30 & 14.71 & -1.33 & 34.93 & -10.13 & 38.81 & -3.08 & 27.36 & -1.81 & 15.83 \\
        & \multicolumn{1}{|l}{PageRank}      & -1.03 & 12.87 & 0.61 & 36.88 & -8.01 & 39.16 & -3.34 & 26.46 & -1.72 & 15.69 \\
        & \multicolumn{1}{|l}{Cluster Coef.} & -0.74 & 12.61 & -0.40 & 24.17 & -5.06 & 42.50 & -3.40 & 24.65 & -1.72 & 12.35 \\
    \midrule
    \multirow{2}{*}{Feature-Based}
        & \multicolumn{1}{|l}{K-Means}        & 0.79 & 11.25 & 0.18 & 25.40 & -2.37 & 41.02 & -4.00 & 23.22 & -2.12 & 12.63 \\
        & \multicolumn{1}{|l}{K-Means (Aggr.)}      & 0.90 & 14.82 & 0.32 & 30.94 & -5.15 & 42.40 & -4.12 & 23.04 & -2.27 & 12.25 \\
    \midrule
    \multirow{3}{*}{Label-Based}
        & \multicolumn{1}{|l}{Neigh-Entropy}        & -0.96 & 13.65 & -3.04 & 30.38 & -0.83 & 45.07 & -3.45 & 24.66 & -1.93 & 12.27 \\
        & \multicolumn{1}{|l}{Max Neigh-Ratio}      & -0.66 & 13.67 & -2.48 & 27.83 & -3.26 & 43.84 & -4.10 & 23.21 & -1.76 & 12.28 \\
        & \multicolumn{1}{|l}{\cellcolor{blue!10}Node Homophily} & \cellcolor{blue!10}-0.62 & \cellcolor{blue!10}21.27 & \cellcolor{blue!10}16.01 & \cellcolor{blue!10}35.83 & \cellcolor{blue!10}18.34 & \cellcolor{blue!10}58.8 & \cellcolor{blue!10}2.97 & \cellcolor{blue!10}41.15 & \cellcolor{blue!10}11.56 & \cellcolor{blue!10}42.74 \\
    \bottomrule
    \end{tabular}
    }
    \label{tab:domain_acc_gain}
    \vspace{-5pt}
\end{table*}

Yet, AMUD implicitly assumes that modeling DPs is beneficial only if these patterns contribute unequally to label prediction. 
This assumption fails in case where all DPs matters similarly important to the label.
Likewise, if all DPs are weakly correlated with the label but one is much stronger than the rest, AMUD may be mistakenly high due to normalization.

To remedy these shortcomings, we propose \textit{Direction Informativeness} (DI), a new metric that directly measures the relative importance of \textit{directed} versus \textit{undirected} patterns for label prediction. Given a digraph $\mathcal{G}_{\text{dir}}$, let $\boldsymbol{d}_{\text{in}}$ and $\boldsymbol{d}_{\text{out}}$ denote the in- and out-degrees. We select the normalized directed degree $\boldsymbol{d}_{\text{dir}} = (\boldsymbol{d}_{\text{in}} - \boldsymbol{d}_{\text{out}})/(\boldsymbol{d}{_\text{in}} + \boldsymbol{d}_{\text{out}})$ to represent the directed pattern, and the bi-directional degree $\boldsymbol{d}_{\text{und}} = \boldsymbol{d}_{\text{in}} + \boldsymbol{d}_{\text{out}}$ as the undirected pattern. Their respective absolute Pearson correlations ($\text{Corr}$) with node labels are:
\begin{equation}
\rho_{\text{dir}} = \left| \mathrm{Corr}(\boldsymbol{d}_{\text{dir}}, \boldsymbol{y}) \right|, \quad
\rho_{\text{und}} = \left| \mathrm{Corr}(\boldsymbol{d}_{\text{und}}, \boldsymbol{y}) \right|,
\end{equation}
where $\boldsymbol{y}$ is the node label vector.
Our Direction Informativeness metric is then defined as:
\begin{equation}
\mathrm{DI}(\mathcal{G}_{\text{dir}}) = \rho_{\text{dir}} \times \log\left( \frac{\rho_{\text{dir}}}{\rho_{\text{und}}} \right),
\end{equation}
where both $\rho_{\text{dir}}$ and $\rho_{\text{und}}$ lie in $[0, 1]$. The logarithmic ratio establishes a clear threshold for when directionality becomes significant (i.e., when $\mathrm{DI} > 0$). The scaling by $\rho_{\text{dir}}$ ensures that cases where both directed and undirected patterns are unrelated to the labels yield low DI (e.g., grid graphs).
We interpret DI values as follows: \textbf{\textit{Informative}:} Values much larger than zero indicate that directionality is highly informative and should be modeled explicitly. \textbf{\textit{Neutral}:} Values positive but close to zero suggest directionality is largely uninformative, yielding only marginal accuracy gains. \textbf{\textit{Adverse:}} Negative values imply that directionality is less informative than simple degree information.

Table~\ref{tab:dir_label_informativeness} summarizes the results, with the experimental setup detailed in Appendix~\ref{app:directional}. Surprisingly, we find that with sufficient hyperparameter and architecture tuning, undirected GNNs often match or even surpass directed GNNs on digraphs. The advantage of directed GNNs, as indicated by previous metrics, appears overstated: only 2 out of 7 datasets show statistically significant improvement over undirected GNNs. In contrast, DI accurately predicts which datasets benefit most from directional modeling, distinguishing clearly between Informative, Neutral, and Adverse regimes.

\begin{table}[ht]
    \vspace{-4pt}
    \centering
    \setlength{\tabcolsep}{4pt}
    \caption{%
    Error Inconsistency (EI) and Complementary Gain (CG) between directed (DirGCN*) and undirected (GCN*) GNNs.
    High EI occurs only on datasets with high Directional Informativeness (DI), but CG remains low, indicating that directed GNNs often correct undirected errors, not vice versa.
    }
    \resizebox{0.49\textwidth}{!}{
    \begin{tabular}{lcccccccc}
    \toprule
    \multirow{2}{*}{Expert Setup} & \multicolumn{2}{c}{\textbf{Arxiv-year}} & \multicolumn{2}{c}{\textbf{Snap-Patents}} & \multicolumn{2}{c}{\textbf{Genius}} & \multicolumn{2}{c}{\textbf{Squirrel-filter}} \\
    \cmidrule(lr){2-3} \cmidrule(lr){4-5} \cmidrule(lr){6-7} \cmidrule(lr){8-9}
    & EI & CG & EI & CG & EI & CG & EI & CG\\
    \midrule
    REINIT         & 13.07 & 6.46 & 11.99 & 5.97 & 2.25 & 1.06 & 15.85 & 7.83 \\
    Dir + UnDir    & 27.92 & 8.41 & 34.13 & 7.87 & 2.79 & 1.26 & 18.89 & 8.18 \\
    \bottomrule
    \end{tabular}
    }
    \label{tab:ensemble_dir_gnn}
    \vspace{-10pt}
\end{table}

We investigate EI and CG between directed and undirected GNNs in Table~\ref{tab:ensemble_dir_gnn}. For datasets (Genius and Squirrel-filter) categorized as \textit{Neutral} by DI, directed GNNs show minimal divergence from undirected GNNs, performing similarly to REINIT. For datasets (Arxiv-year and Snap-Patents) where directionality are \textit{Informative}, combining directed and undirected GNNs produces high error inconsistency but still low complementary gain. This low complementary gain occurs because directed GNNs correctly classify many nodes that undirected GNNs miss, while undirected GNNs rarely correctly predict nodes that directed GNNs fail to classify.

\subsection{Training Data Selection: Domain Discovery on Graphs} \label{sec:train_data_sel}
In this section, we explore strategies for selecting diverse training subsets for expert models, with the goal of maximizing error diversity on the test set. Leveraging advanced graph analytics techniques, we conduct a fine-grained analysis that goes beyond conventional error inconsistency metrics and directly connects the generalization capabilities of experts with specific data characteristics.


We introduce a general and realistic domain setting for graph data, enabling a quantitative evaluation of various training set selection strategies.
Formally, consider the task of predicting node labels $Y\in\mathcal{Y}$ from input features $X\in\mathcal{X}$. We view a graph dataset as a mixture of $M$ domains, denoted as $\mathcal{D}=\{d_1, ..., d_M\}$.
In line with the out-of-distribution (OOD) learning literature~\cite{good}, we define a domain as a node split, where covariate shift in $P(X)$ is present between domains.
For instance, by selecting or constructing a critical feature $X_d$ with continuous and finite values, we can define each domain based on a specific range of $X_d$, leading to a distinct input distribution $P(X)$ for every domain.

For a given domain $d_m$, we denote its input (covariate) distribution as $P_{m}(X)$ and its conditional (concept) distribution as $P_{m}(Y|X)$. The joint distribution over the entire dataset can thus be 
\begin{equation}
P(Y, X) = \sum_{m=1}^{M} w_m P_{m}(Y, X) = \sum_{m=1}^{M} w_m P_{m}(X) P_{m}(Y|X),
\end{equation}
where $w_m$ denotes the mixture proportion for domain $d_m$.
We expect there exist concept shift in $P(Y|X)$ between domains.
We further assume that both the training and test sets are randomly drawn from the same distribution $P(Y, X)$. Specifically, they can be represented as collections of node groups: $V_{\text{train}} = \{\hat{V}_1, \hat{V}_2, ..., \hat{V}_M\}$ and $V_{\text{test}} = \{V_1, V_2, ..., V_M\}$, where each node group $\hat{V}_m$ (or $V_m$) is associated with a particular domain $d_m$.

The standard training procedure for GNN experts relies on empirical risk minimization (ERM) over the entire training set $V_{\text{train}}$. However, this approach can lead to significant generalization failures on specific domains, as observed in many real-world graphs~\cite{demystifying, subgroup}.
To address this issue, we consider a setup in which each expert is trained and tested exclusively on samples from a single domain.
For a given domain $d_m$, the corresponding domain expert---trained on $\hat{V}_m$---is expected to achieve better generalization performance on $V_m$ compared to the ERM baseline trained on the full training set $V_{\text{train}}$.

Building on the above intuition, we propose an evaluation criterion for manual domain assignments.
Consider a domain assignment function $\phi_{\mathrm{dm}}(A, X, Y)$ that outputs an assignment matrix $D \in \{0,1\}^{|V| \times M}$ to assign each node to a domain. Each $i$-th row of $D$ is a one-hot vector indicating the domain assigned to node $v_i$. We evaluate $\phi_{\mathrm{dm}}$ by comparing the performance of domain-specific experts against an expert trained on the complete training set.
Formally, let $\mathrm{Acc}_{m \rightarrow m}$ denote the accuracy of an expert trained on $\hat{V}_m$ and evaluated on $V_m$, while $\mathrm{Acc}_{\mathrm{all} \rightarrow m}$ denotes the accuracy of an expert trained on the full training set and evaluated on $V_m$. Let $w_m$ represent the proportion of nodes in domain $d_m$. The weighted average gain of domain experts $\Delta{\mathrm{Acc}}$ is then defined as
\begin{equation}
\Delta{\mathrm{Acc}} = \sum_{m=1}^M w_m \left(\mathrm{Acc}_{m \rightarrow m} - \mathrm{Acc}_{\mathrm{all} \rightarrow m}\right).\label{eq:acc_gain}
\end{equation}
A higher score indicates better domain assignments. $\Delta{\mathrm{Acc}}$ is a more stringent metric than error inconsistency, as it requires models to achieve better generalization capability while training on fewer samples to obtain a positive score.


In the following, we examine a comprehensive set of heuristic-based domain assignment functions to identify what makes an effective and meaningful domain. We break down the design space of domain assignment function $\phi_{\text{dm}}: A,X,Y \rightarrow D$ into three dimensions: topology-based, feature-based, and label-based. To gain insights into domain discovery for training domain-generalizable experts, we restrict each assignment function to use only one dimension. Table~\ref{tab:domain_acc_gain} collects the overall results.

\begin{figure*}[t]
  \centering
  \vspace{-10pt}
  \includegraphics[width=0.88\textwidth]{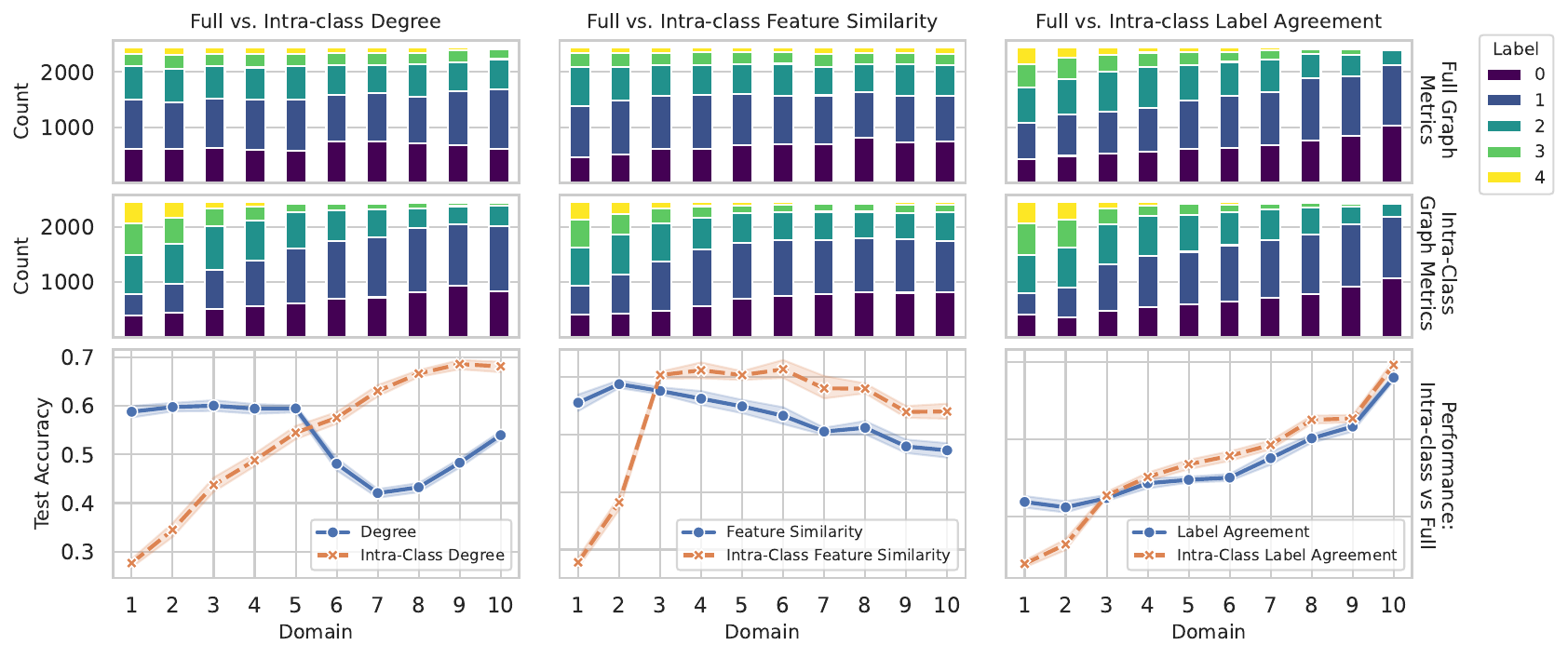}
  \vspace{-10pt}
  \caption{%
    The label distribution and model performance across domains.
    \textbf{Top:} Label distribution (i.e., varied label counts per bin) for full graph metrics. 
    \textbf{Middle:} Label distribution for intra-class graph metrics.
    \textbf{Bottom:} Performance of GCN model (use full training set) across different test domains.
    Domains (node subgroups) partitioned by intra-class metrics (rather than full-graph metrics) exhibit more significant label distribution shifts and greater disparities in expert generalization.
    }
    \vspace{-8pt}
  \label{fig:uniform-label-distribution}
\end{figure*}

\subsubsection{Domain Assignment by Topology}\label{sec:domain-topology}
The topology-only assignment function $\phi_{\text{topo}}: A \rightarrow D$ leverages only the graph topology to assign a domain for each node. We select three widely used node centrality metrics—degree, PageRank, and clustering coefficient—to compute a scalar centrality score for each node. We then evenly split the nodes into two domain groups based on these scores.

The results in Table~\ref{tab:domain_acc_gain} show that \textbf{solely leveraging topology information cannot properly identify domains}, as the accuracy gains $\Delta$Acc are all negative. This suggests that GNN experts fail to achieve better generalization in sparse or dense areas when training samples are restricted to those with similar topological properties.

\subsubsection{Domain Assignment by Feature}
Taking the feature as input, the feature-based assignment function generate node splits for each node $\phi_{\text{feat}}: X \rightarrow D$.
The features of widely used node classification datasets mainly fall into two categories:
(1) tabular data representing node properties (e.g., Snap-Patents~\cite{lim2021large} represents nodes as patents and uses patents' metadata as features)
(2) textual data with word-level embeddings (e.g., arxiv~\cite{ogb} computes word embeddings from each paper's title and abstract as features).
To verify if there exists strong domain indicator in features, we run K-means on raw features to cluster the nodes into to domains. We also include a variant ``K-means (Aggr.)'' that replace raw features with aggregated features $A^2X$.

The results show that feature-based domain assignment can only occasionally achieve limited improvements. This indicates that GNN models trained with standard supervised learning can already effectively capture the mapping between features and labels, even when the feature distribution is diverse.
To further investigate whether certain features can identify distinct domains, we use gradient boosting decision trees to select important features from all available ones. Although we did not find such special patterns in most datasets, we surprisingly discovered that the Genius~\cite{lim2021large} and Question~\cite{critical} datasets contain “magic features” capable of identifying the single-class domain. As shown in Figure ~\ref{fig:special_feat_domain}, by filtering out these trivial samples in domain-1 during training, models can achieve better generalization on the remaining data (domain-0), thereby improving overall performance.

\subsubsection{Domain Assignment by Label}
The label-based domain assignment functions leverage both the topology and label information to identify domains $\phi_{\text{label}}: A, Y \rightarrow D$. 
We seek for popular homophily heuristics to assign domains. To assign each node to a domain, we require node-level homophily metrics instead of more widely known graph-level homophily metrics. We categorize existing node-level homophily metrics into two categories: (1) Neighborhood informativeness, which measures the neighborhood label distribution statistics such as maximum neighborhood label ratio (\cite{zhou2025clarify-588} propose a variant called neighborhood confusion) and neighborhood label entropy. (2) Neighbor Agreement, which calculates the label distribution similarity between the target (self) node and the source (neighboring) nodes, such as the node homophily~\cite{geomgcn}. 
For each selected homophily metric, we first calculate a scalar score for each node and then evenly split the nodes into two domains based on their scores.

\textbf{Neighborhood informativeness metrics prove ineffective, whereas neighbor agreement metric (node homophily) shows strong effectiveness—yielding particularly significant gains on heterophilous datasets.}
Figure~\ref{fig:homophily_domain_performance} illustrates expert accuracy across node subgroups with varying homophily levels. We observe that Expert 0 (EXP=0), trained on low-homophily domains, achieves superior generalization in regions with low homophily but struggles in high-homophily regions. Conversely, Expert 1, trained on high-homophily domains, demonstrates the opposite pattern. This figure highlights an ideal scenario in which the experts are both diverse and complementary.



\subsection{Behind the Scene: Intra-Class Graph Analysis}\label{sec:intra-class}
The above experiments investigate a comprehensive set of domain assignment functions, and we summarize the key findings as follows:
(1) When comparing the EI indicator with baseline REINIT, which does not involve training set selection, we find that training experts on different subsets can effectively diversify expert behaviors---even when the subsets are randomly selected. 
(2) Among all heuristics, only node homophily consistently demonstrates positive gains in $\Delta\text{Acc}$. This highlights the effectiveness and distinctiveness of node homophily, raising a question about what makes a domain assignment effective.

To further understand the effectiveness of node homophily, we revisit its definition:
$\text{h}^{\text{node}}_v = \frac{\left|\left\{u \in \mathcal{N}_v : y_u = y_v \right\}\right|}{|\mathcal{N}_v|}$,
where $\mathcal{N}_v$ denotes the neighbors of node $v$, and $y_v$ is the class label of node $v$. Since node homophily reflects the fraction of neighbors sharing the same class label, it can be reformulated as:
\begin{equation}
    \text{h}^{\text{node}}_v = \frac{\sum_{u \in \mathcal{V}} \mathbbm{1}\{(u, v) \in \mathcal{E}_{\text{intra}}\}}{\sum_{u \in \mathcal{V}} \mathbbm{1}\{(u, v) \in \mathcal{E}\}},
\end{equation}
where $\mathbbm{1}\{\cdot\}$ is the indicator function and $\mathcal{E}_{\text{intra}} = \{(u, v) \in \mathcal{E} \mid y_u = y_v\}$ represents the set of intra-class edges.
This reveals that node homophily is equivalent to the intra-class degree normalized by the full graph degree. Since the full graph degree has been shown ineffective (Section~\ref{sec:domain-topology}), we hypothesize that intra-class structural properties, such as intra-class degree, are critical in determining the quality of domain assignments.

\begin{figure}[t]
  \centering
  \includegraphics[width=0.40\textwidth]{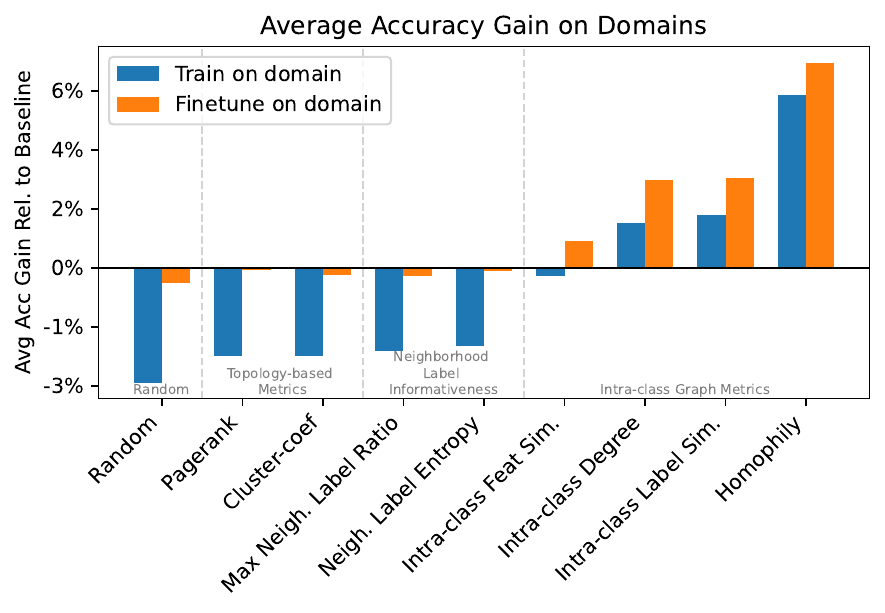}
  \vspace{-10pt}
\caption{%
Impact of domain assignment on average expert accuracy gain, comparing training from scratch vs. fine-tuning on domain-specific data. Gains are computed over a GNN trained on the full graph, averaged across datasets. Intra-class metrics yield the most effective assignments.
}
  \label{fig:bar-rel-acc-improve}
  \vspace{-10pt}
\end{figure}

\noindent\textbf{Intra-Class Graph Metrics.} Formally, we define the intra-class graph as $\mathcal{G}_{\text{intra}}(\mathcal{V}, \mathcal{E}_{\text{intra}}, X, Y)$, where only the original edge set $\mathcal{E}$ is replaced by the intra-class edge set $\mathcal{E}_{\text{intra}}$.
We then introduce a new class of domain assignment functions based on node-level statistics computed on $\mathcal{G}_{\text{intra}}$, which we refer to as intra-class graph metrics.
To align with global metrics, we design three intra-class metrics from the perspectives of topology, features, and labels:

\ding{172}\textit{Intra-class degree}. To capture topological sparsity, we define the intra-class degree as:
    \begin{equation}
        \text{deg}_{v}^{\text{intra}} =\sum_{u\in\mathcal{V}} \mathbbm{1}\{(u,v)\in\mathcal{E}_{\text{intra}}\},
    \end{equation}
which partitions nodes based on their local connectivity in the intra-class graph.

\ding{173}\textit{intra-class feature similarity}. To leverage feature information, we compute the average feature similarity between a node and its intra-class neighbors using the dot product:
    \begin{equation}
        \text{feat}_{v}^{\text{intra}} = \frac{1}{\text{deg}_{v}^{\text{intra}}} \sum_{u\in\mathcal{V}} \mathbbm{1}\{(u,v)\in\mathcal{E}_{\text{intra}}\} \langle \mathbf{x}_u, \mathbf{x}_v \rangle.
    \end{equation}
A higher value indicates that neighbors are more semantically aligned with their labeling.

\ding{174}\textit{intra-class label agreement}. To assess label-level consistency, we first compute the neighborhood label distribution using the full graph: $\tilde{\mathbf{y}}_v = \sum_{u\in\mathcal{V}} \mathbbm{1}\{(u,v)\in\mathcal{E}\} \mathbf{y}_u$, where $\mathbf{y}_v$ is the one-hot label vector for node $v$. Then, we define the label agreement between a node and its intra-class neighbors as:
    \begin{equation}
        \text{label}_{v}^{\text{intra}} = \frac{1}{\text{deg}_{v}^{\text{intra}}} \sum_{u\in\mathcal{V}} \mathbbm{1}\{(u,v)\in\mathcal{E}_{\text{intra}}\} \langle \tilde{\mathbf{y}}_u, \tilde{\mathbf{y}}_v \rangle.
    \end{equation}
This metric captures how similar the neighborhood label distributions are between intra-class connected nodes.
Finally, we classify node homophily as a special case of intra-class graph metrics, computed as $\text{h}^{\text{node}}_v = \text{deg}^{\text{intra}}_v / \text{deg}_v$



To test our hypothesis, we assess the effectiveness of intra-class metrics by directly comparing them to their global counterparts. The global variants (e.g., $\text{deg}_v$, $\text{feat}_v$, and $\text{label}_v$) are computed by substituting the intra-class edge set $\mathcal{E}_{\text{intra}}$ with the full edge set $\mathcal{E}$. We first train a GCN model on the entire training set, and then measure its accuracy across different test domains defined by either intra-class or global metrics. As shown in Figure ~\ref{fig:uniform-label-distribution} (top and middle rows), domains identified by intra-class metrics exhibit more pronounced concept shifts. The bottom row further demonstrates that GNNs show greater performance disparities across intra-class domains.
We further investigate an expert training approach that first pre-trains experts on the full training data, and then fine-tunes them on their corresponding target domain. Figure ~\ref{fig:bar-rel-acc-improve} highlights the effectiveness of this approach.

\begin{figure}[t]
 \vspace{-10pt}
  \centering
  \includegraphics[width=0.45\textwidth]{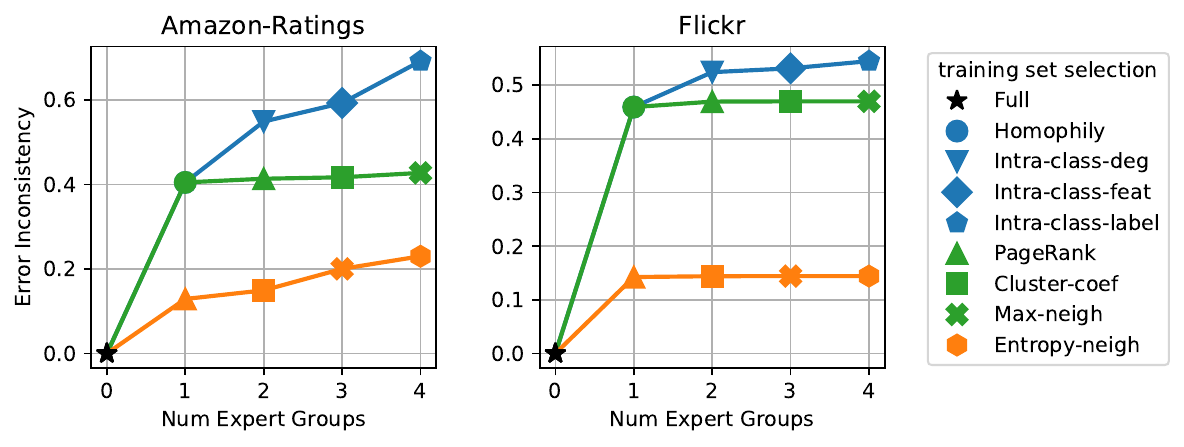}
  \caption{%
     Relative EI improvements of GCN ensembles, where experts are trained on full data and fine-tuned on nodes selected by domain assignment functions. We show the ensemble performance of ensembling \textcolor{darkblue}{intra-class expert groups}, \textcolor{darkgreen!90}{homophily + others}, and \textcolor{orange}{without intra-class expert groups}. Intra-class experts are shown to be critical in improving ensemble performance.
  }
  \vspace{-10pt}
  \label{fig:line-ei-ensemble}
\end{figure}

\begin{figure}[t]
  \centering
  \includegraphics[width=0.48\textwidth]{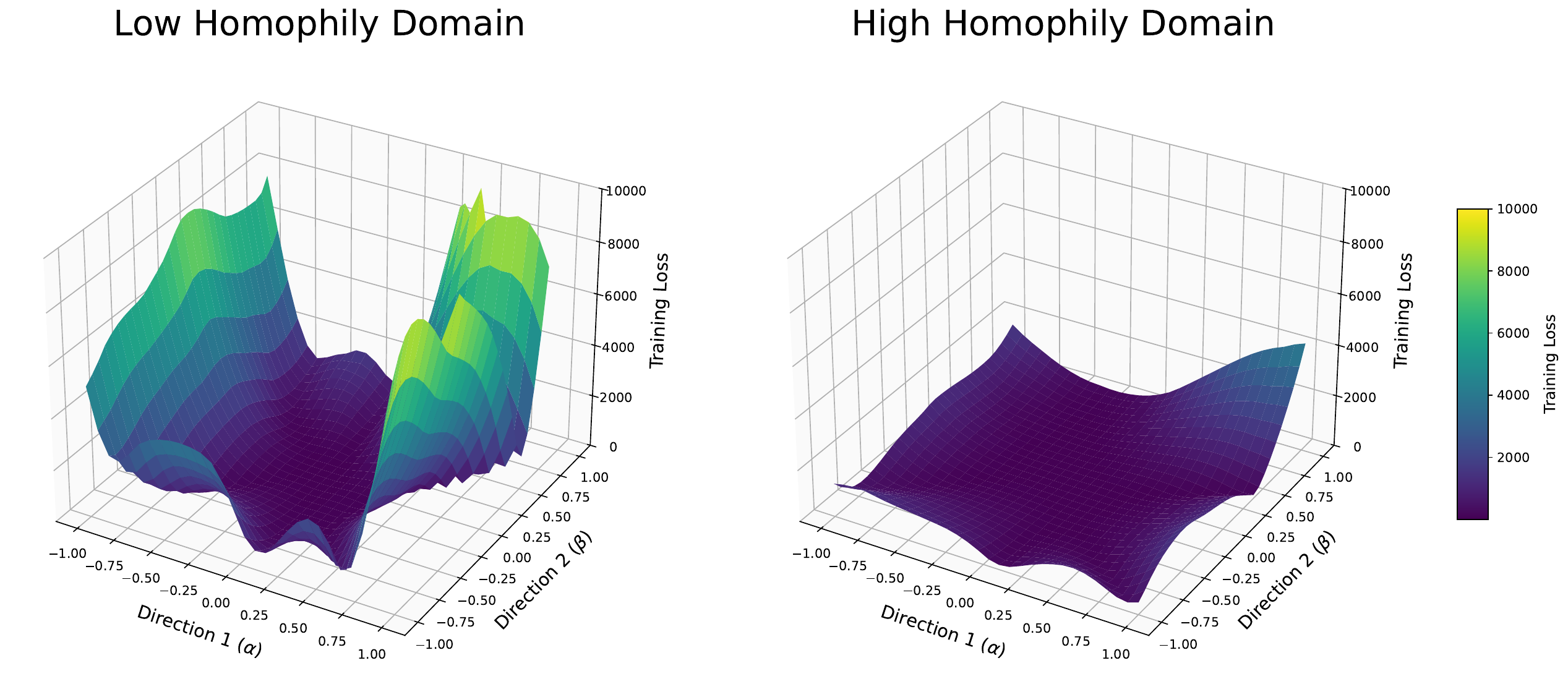}
  \caption{%
    \textbf{GCN loss landscapes on Penn94.} (Left) Expert trained on low-homophily nodes shows a rugged surface with many sharp minima, indicating difficult optimization. (Right) High-homophily expert has a smoother, convex landscape, suggesting an easier training task.
}
  \label{fig:loss_landscape}
  \vspace{-10pt}
\end{figure}

\noindent\textbf{Interpretation and Visualization.} To further interpret why intra-class graph metrics are effective, we analyze the training difficulty of the domains they define. 
We hypothesize that metrics like node homophily serve as strong indicators of training difficulty, effectively separating the data into “easy” and “hard” domains. The presence of such clearly distinct domains provides strong motivation for training specialized experts, as a single model may struggle to generalize across them---risking underfitting on hard domains and overfitting on easy ones.
We empirically validate this hypothesis by showing the loss landscape across intra-class metrics in Figure \ref{fig:loss_landscape}. 
Figure ~\ref{fig:loss_landscape} demonstrates that intra-class metrics like homophily correspond to training difficulty. We observe that the high-homophily expert’s landscape is smooth and convex, indicating easier optimization. In contrast, the low-homophily expert’s landscape is rugged, with many sharp minima, signaling a much more challenging training process. We further presents t-SNE embeddings on the Flickr dataset in Figure ~\ref{fig:tsne_domain_comparison}, showing an improved separation brought by intra-class metric (see Appendix ~\ref{app:vis-intra} for details).

\subsection{How Do Diversification Techniques Benefit Downstream Task Performance?}\label{sec:downstream}

In this section, we assess the effect of various expert diversification techniques on downstream node classification performance by aggregating the diversified experts with Global Ensemble (Ensemble), Moscat Gating Ensemble (MoE), and an oracle gating model (Table ~\ref{tab:downstream-task}). We highlight several key findings:
\textbf{(1) Performance Gains from Diversification:} Both Ensemble and MoE significantly improve upon well-tuned single GNN models when appropriate diversification techniques are used. Notably, employing a single diversification strategy with just three experts can yield up to a 4\% performance gain over the best individual model. We anticipate that further gains are possible by incorporating more experts, combining multiple diversification strategies, and expanding the range of model architectures.
\textbf{(2) Effect of Expert Number ($K$):} Increasing the number of experts via a single diversification technique yields diminishing improvements in performance. Our results focus on $K \leq 3$, as limited additional benefit is observed for $K \geq 4$. For data partitioning-based diversification, increasing $K$ reduces the domain size assigned to each expert, which can lead to higher specialization but smaller training sets. For instance, on Amazon-Ratings, increasing $K$ from 2 to 3 in the “Homophily” setting resulted in decreased Ensemble/MoE accuracy but increased upper-bound accuracy. Fine-tuning pre-trained experts on their target domains can help balance diversity, complementarity, and overall ensemble accuracy.
\textbf{(3) Exploiting Weak Experts:} While diversification reliably produces a variety of experts, it may sometimes yield weak experts with low individual accuracy, potentially hindering ensemble gains. However, MoE typically mitigates this issue by leveraging expert diversity more effectively through adaptive gating. We hypothesize that MoE can better calibrate over- or under-confident experts in specific regions, whereas a simple Ensemble cannot.

\section{Conclusion}




This work systematically investigated the importance and strategies for expert diversification in Graph Neural Network Mixture-of-Experts (GNN-MoE) frameworks. We analyze a wide spectrum of diversification techniques across standard node classification benchmarks through the concepts of Error Inconsistency (EI), Complementarity Gain (CG), and ensemble accuracy, drawing a strong connection between expert diversity and ensemble accuracy across both models and datasets.
We organize our study around five major diversification strategies and highlight three central findings:

\noindent \textbf{Training data selection is the most useful diversification strategy.} Among all strategies, training data selection, especially using domain assignment functions based on intra-class metrics, yields the greatest gains in expert diversity and ensemble performance. We both empirically validate and theoretically analyze the benefits of intra-class metrics, which better capture localized graph properties than full-graph metrics.

\noindent \textbf{Directional Modeling has limited effectiveness.}
While directionality can increase diversity, it does not consistently improve ensemble performance. We propose a new metric, Directional Informativeness (DI), to identify when edge directions are beneficial in diversification. Our analysis further shows that diversity from directed GNNs may not translate into accuracy gains.

\noindent \textbf{Effective expert diversification leads to substantial downstream performance improvements.} Across two ensemble strategies, we observe that strong diversification contributes to accuracy gains, affirming the practical value of our comprehensive diversification analysis.

\bibliographystyle{ACM-Reference-Format}
\bibliography{cite}

\clearpage
\appendix
\section{Related Work}\label{sec:relate}

\subsection{Diversity in Model Ensemble and Soup}

Ensembling, which combines the outputs of multiple models, is a foundational technique for improving the accuracy and robustness of machine learning models~\cite{bauer1999empirical, opitz1999popular}. 
To reduce the additional computation and memory costs brought by inferencing with more models, model souping is proposed. By averaging the weights of multiple separately trained models, model souping only uses one model during inference~\cite{kolesnikov2020bigtransferbitgeneral, wortsman2022model-e06, zuber2025enhanced-4b1}. However, model souping comes with the limitation that all models must share the same architecture, and its performance generally lags behind ensembling~\cite{wortsman2022model-e06}.

Existing study have empirically showed that ensemble performance is correlated with model soup performance~\cite{wortsman2022model-e06}, and diversity among models is the key to achieving strong results with both techniques~\cite{wen2020batchensemblealternativeapproachefficient, no-one, oshana2022efficient-cc3}. Specifically, Gontijo-Lopes et al.~\cite{no-one} conducted a large-scale empirical study evaluating various training methodologies to promote model diversity. Wortsman et al.~\cite{wortsman2022model-e06} explored averaging models trained with varied hyperparameter configurations to improve model souping performance.
In graph learning, adapting ensembling and souping has also attracted substantial interest, with numerous diversification techniques proposed for model mixing. For instance, Oshana et al.
~\cite{oshana2022efficient-cc3} leverage network and graph regularization to diversify models; Zuber et al.~\cite{zuber2025enhanced-4b1} propose using METIS partitioning to generate subgraphs, training each model on a different subgraph for model souping; and Wei et al.~\cite{wei2023gnnensemblerandomdecisiongraph} diversify models by randomly selecting substructures in the topological space and subfeatures in the feature space.



\subsection{Mixture of Experts on Graphs} 
Mixture-of-Experts (MoE) frameworks have recently gained significant attention in graph learning, leading to a surge of novel approaches and applications. An early work~\cite{graphmoe} referencing the LLM MoE introduced top-$K$ sparse gating techniques to effectively combine multiple GNN experts.
MoE frameworks have since been adopted for a wide range of graph learning tasks. For instance, GraphMetro~\cite{graphmetro} leverages MoE models to detect complex distribution shifts in out-of-distribution scenarios. Node-MoE~\cite{han2024node-wise-34e} applies MoE strategies to adaptively select filters tailored to individual nodes for classification, while Link-MoE~\cite{ma2024mixture-d0d} dynamically assigns the most suitable expert to each node pair based on their pairwise properties for link prediction tasks.

More recent work has focused on promoting diversity among experts to further enhance MoE performance. Mowst~\cite{mowst} proposed separating self-features from neighborhood information by mixing weak MLP experts with strong GNN experts. Chen et al.~\cite{chen2025mixturedecoupledmessagepassing} construct diverse message-passing experts by recombining fine-grained encoding operators. Moscat~\cite{moscat} proposes decoupling the training of the gating model from the experts and demonstrates more accurate capture of experts' generalization behaviors.
\section{Experimental Setups}

\subsection{Datasets}\label{app:datasets}
To showcase model behavior on real-world datasets exhibiting diverse and complex patterns, we include both homophilous datasets (datasets with high homophily ratios) and non-homophilous datasets spanning a range of homophily levels. We also ensure coverage of datasets with different scales to provide a comprehensive evaluation across small ($<$1k nodes), medium (1k$\sim$100k nodes), and large graphs ($>$100k nodes). Importantly, we exclude widely used small heterophilous graphs (with only a few hundred nodes), as their high variance undermines statistical significance~\cite{critical}. We also omit datasets with synthetically generated edges, such as Minesweeper and Roman-Empire~\cite{critical}.
Based on these criteria, we select the following 14 datasets:
(1) Citation networks: {Pubmed}~\cite{Planetoid}, {Ogbn-Arxiv}~\cite{ogb}, {Arxiv-year}~\cite{lim2021large}, and {Snap-patents}~\cite{lim2021large};
(2) Wiki networks: {Chameleon-filter}~\cite{critical}, {Squirrel-filter}~\cite{critical}, {Actor}~\cite{geomgcn}, and {WikiCS}~\cite{wikics};
(3) Co-purchase networks: {Computer}~\cite{shchur2019pitfallsgraphneuralnetwork}, {Amazon-Ratings}~\cite{critical};
(4) Co-occurrence networks: {Flickr}~\cite{saint};
(5) Social networks: {Penn94}~\cite{lim2021large}, {Genius}~\cite{lim2021large}, and {Questions}~\cite{critical}.
For {Pubmed}, {Actor}, and {Computer}, we adopt the widely used 60\%/20\%/20\% random splits for training, validation, and testing, following~\cite{acmgcn}. For all other datasets, we use the original data splits.

\subsection{Expert Architectures}\label{app:expert_arch}
We restrict the architecture of our experts to Message-Passing GNNs, leaving other neural network architectures such as Graph Embeddings or Graph Transformers for future work. Within this space, we integrate a variety of GNN techniques into a unified modular framework, where each approach corresponds to a specific architectural component. For instance, the graph filter component includes implementations from GCN~\cite{gcn}, MixHop~\cite{mixhop}, and DirGCN~\cite{dirgnn}; for skip connections, we incorporate GraphSAGE-style concatenation~\cite{sage}, Jumping Knowledge networks~\cite{jknet}, and so on. The framework also includes standard layers such as normalization and dropout.
To isolate the impact of each component, we vary one component while fixing the remaining components as hyperparameters, thereby fully illustrating the role of the selected component.

\section{Expert Diversification} \label{app:arch-param-ei}

\subsection{Extended Results for Diversification Strategy Overview}

\begin{figure*}[!htbp]
\vspace{-5pt}
  \centering
  \includegraphics[width=0.7\textwidth]{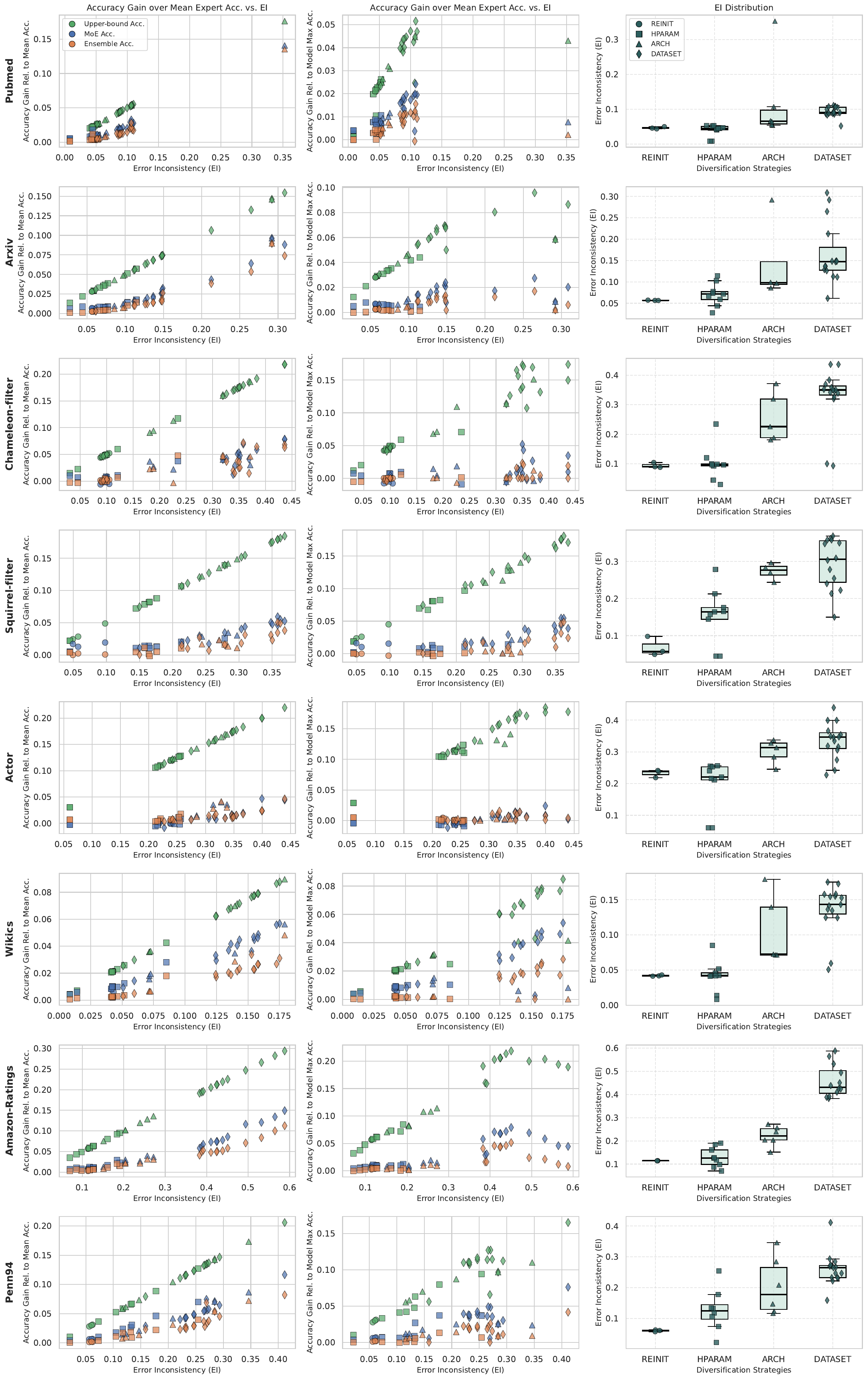}
  \vspace{-8pt}
  \caption{%
    Overview of the effectiveness of diversification strategies across various datasets. Each datapoint corresponds to a pair of GNNs diversified using a specific strategy. 
    \textbf{(left)} The relationship between accuracy gain (relative to the mean expert accuracy) and error inconsistency (EI). 
    \textbf{(center)} The relationship between accuracy gain (relative to the max expert accuracy) and EI.
    \textbf{(right)} Boxplots of EI's distribution across the four expert diversification strategy categories.
  }
  \label{fig:multi-dataseti-ei-full}
\end{figure*}

Figure~\ref{fig:multi-dataseti-ei-full} demonstrates an overview of the effectiveness of various diversification strategies, in terms of diversity (EI), complementarity (Upper Bound Acc), and ensemble/MoE accuracy, across different categories and datasets. We differentiate categories with different shapes: REINIT (circal), HPARAM (square), ARCH (triangle), and DATASET (diamond), and differentiate expert ensemble approaches in different colors: \textcolor{darkgreen!90}{\emph{Upper Bound}}, \textcolor{darkblue}{\emph{Moscat Gating Ensemble} (MoE)}, and \textcolor{orange}{\emph{Global-Ensemble} (Ensemble)}. The left and center columns show the accuracy gain of expert ensembles over mean expert accuracy and max expert accuracy, respectively. 




\subsection{Extended Results for Hyperparameter Search for Divergent Experts}\label{app:param-ei}

\begin{figure*}[!htbp]
\vspace{-5pt}
  \centering
  \includegraphics[width=0.75\textwidth]{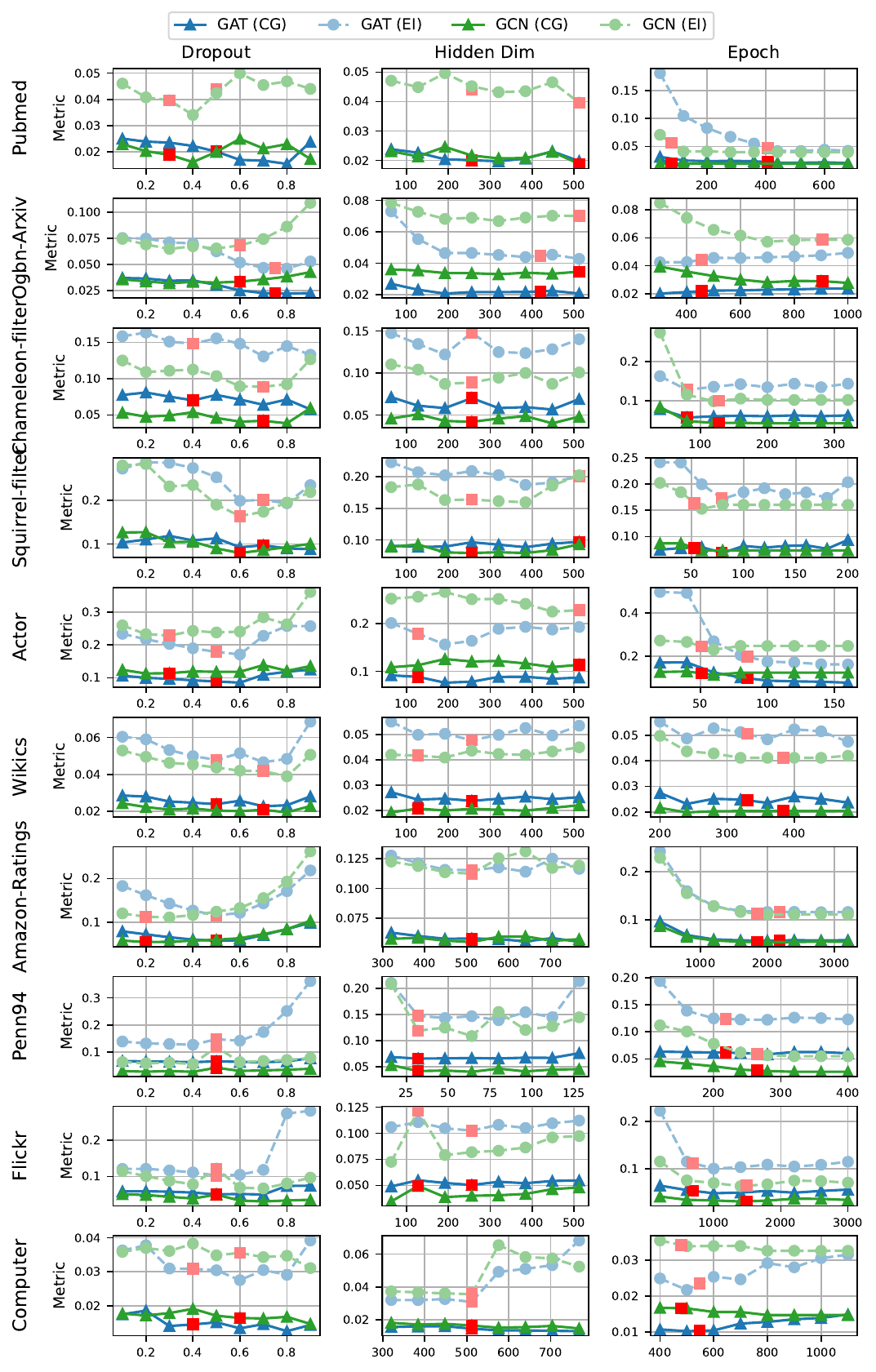}
  \vspace{-8pt}
  \caption{%
    Full pairwise Error Inconsistency (EI) and Complementary Gain (CG) trnd with a single modified hyperparameter.  %
    For every architecture (GCN or GAT), we first get the \emph{best-performing expert} via hyperparameter tuning and use its configuration as the baseline.  %
    We then create a \emph{single-hyperparameter variant} and measure their inconsistency with respect to the baseline model (baseline values marked by the red squares). %
    Each point shows the average inconsistency over five random seeds.  %
    \textbf{(left)} varying dropout ratio (0.1–0.9).  %
    \textbf{(center)} varying hidden dimension (64–768; dataset-dependent).  %
    \textbf{(right)} varying training epochs (0–3600; dataset-dependent).  %
  }
  \label{fig:hparam-inconsistency-full}
\end{figure*}

To systematically evaluate expert diversification, we conduct a controlled search over general hyperparameters and architectural components. For each experiment, we begin by tuning a base GNN (either GCN or GAT) for accuracy using established protocols~\cite{tuned}. We then isolate a single design factor (dropout rate, hidden dimension, or number of layers) and introduce a perturbed model. The perturbed model is compared to the base model by computing their pairwise error inconsistency (EI) and complementary gain (CG), measuring the degree of behavioral divergence induced by the perturbation.

Figure~\ref{fig:hparam-inconsistency-full} reports error inconsistency under variations in dropout rate, hidden dimension, and training epochs.
We focus on the dropout rate, hidden dimension, and training epochs as they are widely tuned and important hyperparameters that affect model regularization, capacity, and convergence behavior. 
We find that even simple hyperparameter changes can produce substantial EI, particularly when deviating far from the base setting. While EI generally correlates with complementary gain (CG), higher EI does not always imply higher CG. 
For instance, on the Squirrel dataset, models trained with fewer epochs exhibit large EI but low CG, likely due to underfitting—yielding predictions that differ from the baseline but lack useful signal.
Across datasets, we often observe a U-shaped pattern: EI and CG are lowest near the tuned setting (red square) and increases with greater deviation. For example, on Amazon-Ratings, changing dropout from 0.5 to either 0.25 or 0.75 nearly doubles EI. This implies that even common tuning techniques can be surprisingly effective for expert diversification.
The impact of hyperparameter perturbations is also dataset-dependent; for example, Actor and Squirrel are notably more sensitive than others.

\subsection{Extended Results for Architectural Search for Divergent Experts}\label{app:arch-ei}

\begin{figure*}[!htbp]
\vspace{-5pt}
  \centering
  \includegraphics[width=0.75\textwidth]{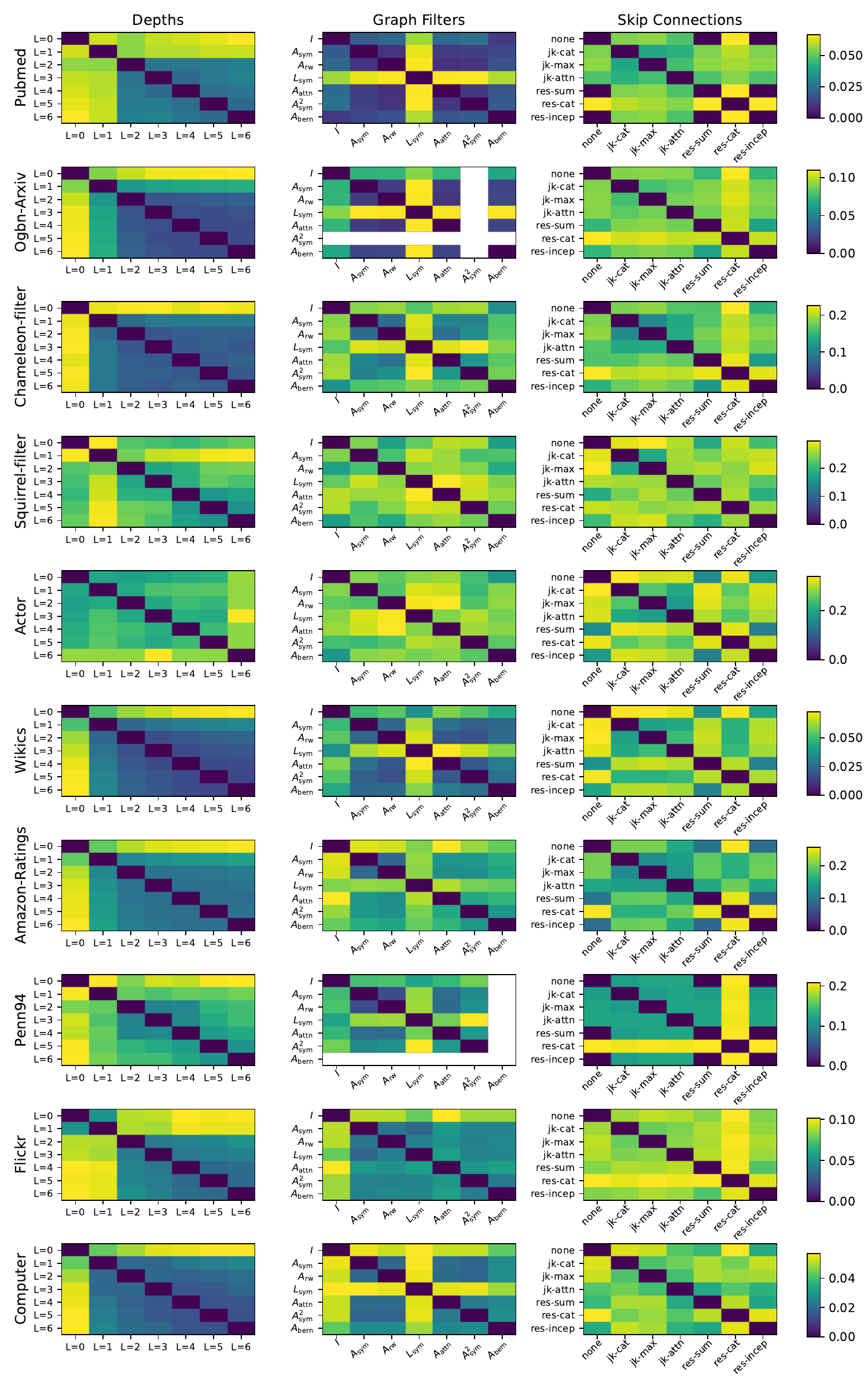}
  \vspace{-8pt}
  \caption{%
    Full pairwise error inconsistency across architectural design dimensions. Each heatmap shows the average error inconsistency (computed over 10 runs) between expert pairs differing on a single design choice. \textbf{(left)} GNN depth ($L{=}0$-$6$),
   \textbf{(center)} graph filters,
   and \textbf{(right)} skip-connections. Each row corresponds to one of ten datasets, and all subplots share a common color scale (shown on the right). Blank entries in the center column for arxiv and penn94 indicate missing data due to out‐of‐memory errors.%
  }
  \label{fig:arch-inconsistency-full}
\end{figure*}

We display the complete experimental results of Error Inconsistency (EI) with different architectural variations of GNNs in Figure ~\ref{fig:arch-inconsistency-full}. For depth (left), the shallowest model ($L{=}0$) consistently exhibits the highest EI, reflecting its fundamentally different behavior due to a lack of message passing. On the other hand, neighboring depths (e.g., $L=3$ vs.\ $L=4$) show smaller inconsistencies. For graph filters (center), we use GATConv as a learnable attention-based graph filter ($A_{\text{attn}}$), and use BernConv defined in ~\cite{he2021bernnet} as a polynomial filter ($A_{\text{bern}}$). We observe that non-smoothing operators, such as the identity matrix ($I$) and high-pass filters like the Laplacian ($L_{\text{sym}}$), often result in high EI, especially on homophilic datasets like Pubmed and Wikics. Finally, skip connection strategies (right) introduce more modest differences; we include both residual and jumping knowledge (JK) variants: jk-max selects the maximum activation across layers, jk-attn learns attention weights over layers, res-cat concatenates layer outputs as in GraphSAGE-style residuals, and res-incep performs residual fusion with inception-style compression. These variants yield dataset-dependent EI patterns.

\subsection{Evaluation of Directional Modeling}\label{app:directional}
\begin{table*}[t]
    \centering
    \caption{
    Comparison of directional modeling metrics on digraph benchmarks. Our proposed \textbf{Direction Informativeness (DI)} metric best aligns with actual accuracy differences, correctly identifying only 2 out of 7 datasets as truly benefiting from directionality(see blue and green highlights).
    }
    \resizebox{0.85\textwidth}{!}{
    \begin{tabular}{llccccccccc}
    \toprule
    \multirow{2}{*}{\textbf{Categories}} & \multirow{2}{*}{\textbf{Datasets}} & \multicolumn{4}{c}{\textbf{Directional Modeling Metrics}} & \multicolumn{2}{c}{\textbf{Undirected GNNs}} & \multicolumn{2}{c}{\textbf{Directed GNNs}} \\
    \cmidrule(lr){3-6} \cmidrule(lr){7-8} \cmidrule(lr){9-10}
    & & $\text{h}_{\text{edge}}$ & $\text{h}_{\text{node}}$ & AMUD & \textbf{DI (Ours)} & GCN* & MixHop* & DirGCN* & HoloNet* \\
    \midrule

    \multirow{2}{*}{Informative} & \multicolumn{1}{|l|}{Arxiv-Year~\cite{lim2021large}}   & 0.22 (D-) & 0.28 (D-) & 0.63 (D-)  & \multicolumn{1}{c||}{\cellcolor{DIblue3} 3.24 (D-)}  & 54.29 & 55.27 & \cellcolor{Green2.5}65.78 & \cellcolor{Green2.5}66.04  \\
                                 & \multicolumn{1}{|l|}{Snap-Patents~\cite{lim2021large}} & 0.07 (D-) & 0.19 (D-) & 0.59 (D-)  & \multicolumn{1}{c||}{\cellcolor{DIblue3} 0.88 (D-)}  & 54.96 & 55.05 & \cellcolor{Green2.5}73.95 & \cellcolor{Green2.5}75.10 \\
    \midrule
    \multirow{2}{*}{Neutral}     & \multicolumn{1}{|l|}{Squirrel-filter~\cite{critical}}  & 0.21 (D-) & 0.16 (D-) & 0.66 (D-) & \multicolumn{1}{c||}{\cellcolor{DIblue5} 0.05 (U-)}  & 44.79 & 43.49 & \cellcolor{Green1}45.36 & \cellcolor{Green1}44.83 \\
                                 & \multicolumn{1}{|l|}{Genius~\cite{lim2021large}}       & 0.62 (U-) & 0.51 (U-) & 0.71 (D-) & \multicolumn{1}{c||}{\cellcolor{DIblue5} 0.05 (U-)}  & \cellcolor{Green1}92.32 & \cellcolor{Green1}92.34 & 92.28 & 92.29 \\           
    \midrule
    \multirow{3}{*}{Adverse}     & \multicolumn{1}{|l|}{Ogbn-Arxiv~\cite{ogb}}            & 0.66 (U-) & 0.64 (U-) & 0.47 (U-) & \multicolumn{1}{c||}{\cellcolor{DIblue5} -0.00 (U-)} & \cellcolor{Green1.5}73.29 & \cellcolor{Green1.5}73.33 & 72.55 & 72.63 \\
                                 & \multicolumn{1}{|l|}{Actor~\cite{geomgcn}}             & 0.22 (D-) & 0.21 (D-) & 0.36 (U-) & \multicolumn{1}{c||}{\cellcolor{DIblue5} -0.00 (U-)} & \cellcolor{Green1.5}43.35 & \cellcolor{Green1.5}42.30 & 42.16 & 41.84 \\
                                 & \multicolumn{1}{|l|}{Chameleon-filter~\cite{critical}} & 0.24 (D-) & 0.20 (D-) & 0.70 (D-) & \multicolumn{1}{c||}{\cellcolor{DIblue5} -0.04 (U-)} & \cellcolor{Green1.5}47.01 & \cellcolor{Green1.5}47.23 & 45.20 & 45.99 \\
    
    \bottomrule
    \end{tabular}
    }
    \label{tab:dir_label_informativeness}
\end{table*}

To evaluate the effectiveness of directional modeling, we compare the accuracy of undirected and directed GNNs on several benchmark digraph datasets. For undirected GNNs, each digraph is converted to its undirected form, while for directed GNNs, the original structure is preserved. To ensure a fair comparison, we select a specific graph filter for each variant and tune all other architectural components and hyperparameters within a unified search space (see Section~\ref{sec:arch-search}). Specifically, we use GCN (first-order) and MixHop (second-order) as undirected graph filters, and DirGCN (first-order) and HoloNet (second-order) as directed graph filters. To distinguish from their original implementations, we mark each variant with an asterisk (*).

\section{Results for Domain Discovery} \label{app:domain-disc}
\subsection{Domain Assignment by Feature}
\begin{figure}[H]
  \centering
  \hspace*{0.4cm}
  \includegraphics[width=0.40\textwidth]{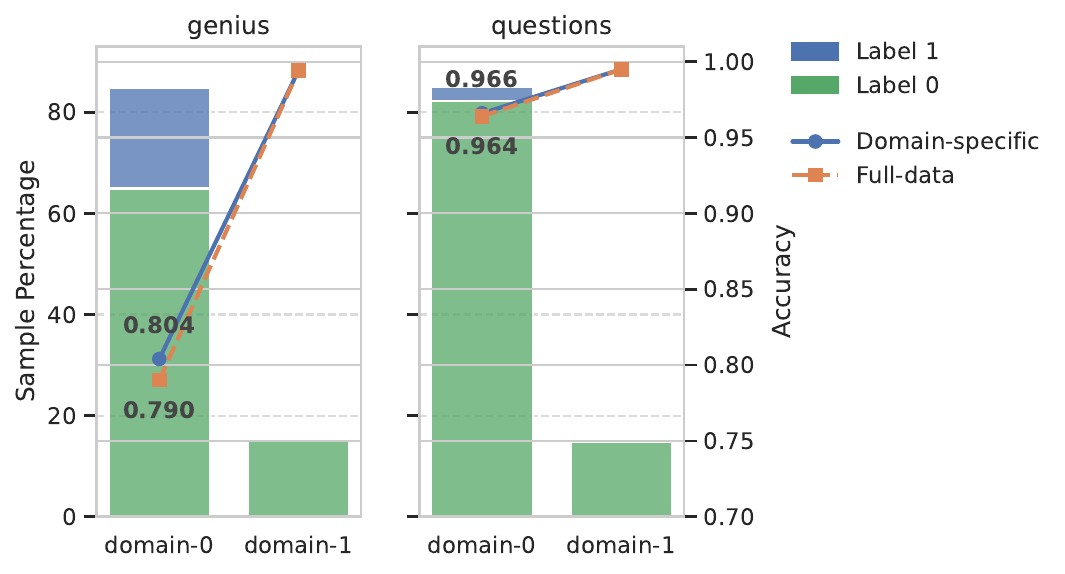}
  \caption{%
  Label imbalance and accuracy in the Genius and Questions datasets. 
  \texttt{Domain-0} is skewed toward \texttt{Label 1}, while \texttt{Domain-1} favors \texttt{Label 0}. 
  Solid blue: domain-specific experts; dashed orange: full expert. 
  Domain experts outperform despite limited data access.
}
  \label{fig:special_feat_domain}
  \vspace{-8pt}
\end{figure}

We discovered that the Genius~\cite{lim2021large} and Question~\cite{critical} datasets contain “magic features” capable of identifying the single-class domain (Figure ~\ref{fig:special_feat_domain}).

\subsection{Domain Assignment by Label}
\begin{figure}[H]
\vspace{-10pt}
  \centering
  \includegraphics[width=0.40\textwidth]{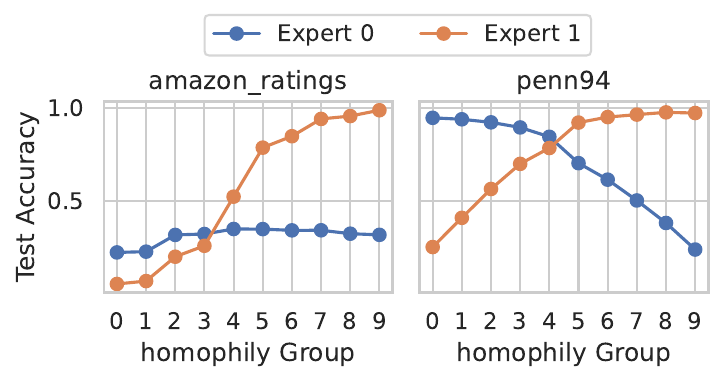}
  \caption{%
  GAT expert performance across homophily-based groups on Amazon-Ratings and Penn94. 
  Groups are ordered by homophily (0: most heterophilic, 9: most homophilic). 
  Experts are trained on the bottom or top 50\% of nodes and evaluated on all groups.
}
  \label{fig:homophily_domain_performance}
  \vspace{-10pt}
\end{figure}

Figure \ref{fig:homophily_domain_performance} reports expert accuracy after grouping nodes by their homophily score. We observe that domain assignment by label revealing neighbor-agreement (node homophily) is highly predictive in aligning experts with domains.

\subsection{Interpretation and Visualization of Intra-class Metrics} \label{app:vis-intra}
\begin{figure*}[t]
\vspace{-4pt}
  \centering
  \includegraphics[width=0.7\textwidth]{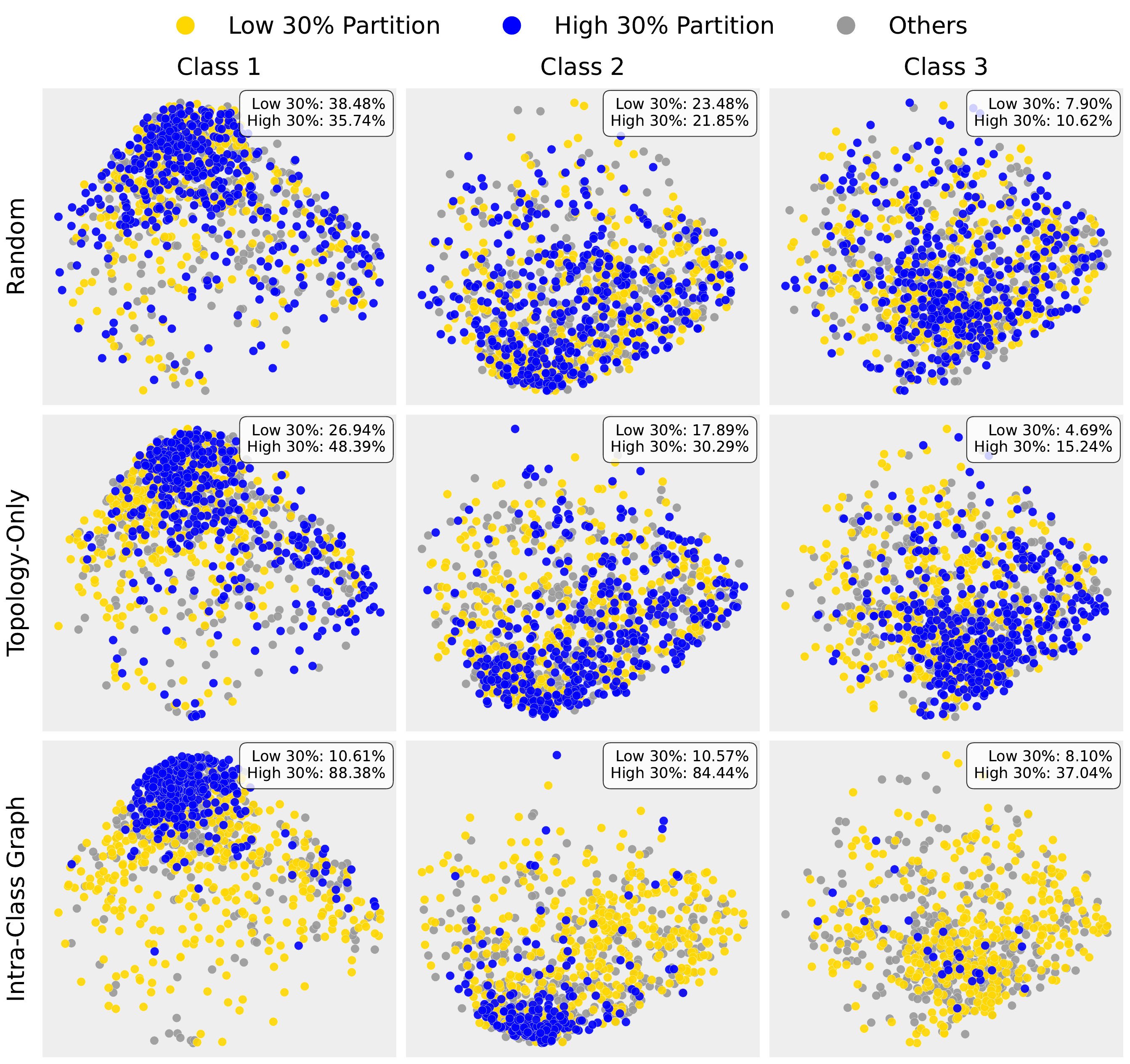}
  \caption{%
    T-SNE visualization of node embeddings trained on the entire dataset using three different domain assignment strategies. 
  \textbf{Top:} Random assignment yields overlapping clusters. 
  \textbf{Middle:} Topology-based metric shows mild separation. 
  \textbf{Bottom:} Intra-class metric (homophily) results in more distinct clusters. 
  Nodes are colored by 30/40/30 percentile groups; accuracy is reported per partition.
}
  \label{fig:tsne_domain_comparison}
  \vspace{-8pt}
\end{figure*}

Figure ~\ref{fig:tsne_domain_comparison} presents t-SNE embeddings on the Flickr dataset of selected classes, where nodes are partitioned by different metrics. Only the intra-class metric produces clear spatial separation: high-value nodes (the easy domain) cluster densely at the core, while low-value nodes (the hard domain) form a sparse outer halo. This structural split is mirrored in model performance, as the annotated accuracies consistently reveal a substantial gap between hard and easy domains.

\subsection{Discussion of Domain Setup}
We compare our domain setup with related literature to provide the readers with a clear positioning of the work. One line of research~\cite{good, graphmetro} addresses the out-of-distribution (OOD) challenge on graphs. In the OOD setting, the training data consists of several domains, but the test domains may not overlap with those seen during training. In contrast, our setting assumes that both the training and test sets are drawn from the same distribution. Specifically, OOD learning aims to capture domain-invariant patterns so that models generalize well to unseen domains, whereas our MoE approach seeks to specialize experts for each domain and employs a gating model to adaptively combine their outputs.
Another line of work~\cite{moscat,demystifying,subgroup} examines the generalization performance of GNNs across different test domains. However, these studies typically train the model on the entire training set, with limited investigation into how the selection of training nodes impacts performance.

\section{Results for Downstream Tasks} \label{app:downstream}

\begin{table*}[!ht]
    \centering
    \caption{
     Accuracy comparison of node classification using Global-Ensemble (Ensemble) and Moscat Gating Ensemble (MoE). We also include the Upper-bound Accuracy to indicate the oracle ensemble of experts. 
     "-FT" denotes experts that were pretrained on the full training set and then individually finetuned on their respective target domains. Notably, \colorbox{gray!10}{"Homophily(-FT)"} uses test labels to calculate homophily ratios (not practical in real scenarios). The ``Filtered-Homophily-FT'' avoids this issue by excluding test labels when calculating training set homophily ratios.
    }
    \resizebox{0.8\textwidth}{!}{
    \begin{tabular}{l|ccc|ccc}
    \toprule
    \multirow{2}{*}{\textbf{Method} \textbf{(Num. Experts)}} 
    & \multicolumn{3}{c|}{\textbf{Amazon-Ratings}} 
    & \multicolumn{3}{c}{\textbf{Penn94}} \\
    \cmidrule(lr){2-4} \cmidrule(lr){5-7}
    & Ensemble & MoE & Upper Bound
    &  Ensemble & MoE & Upper Bound \\
    \midrule
    \multicolumn{1}{l|}{GCN \textbf{(1)}}          & 54.10 \std{0.55} & 54.10 \std{0.55} & 54.10 \std{0.55}  & 82.46 \std{0.43} & 82.46 \std{0.43} & 82.46 \std{0.43}\\
    \multicolumn{1}{l|}{REINIT \textbf{(2)}}       & 54.70 \std{0.33} & 55.00 \std{0.36} & 59.79 \std{0.54}  & 82.63 \std{0.43} & 83.15 \std{0.51} & 85.48 \std{0.56} \\
    \multicolumn{1}{l|}{REINIT \textbf{(3)}}       & 54.67 \std{0.34} & 55.03 \std{0.35} & 62.64 \std{0.54}  & 82.72 \std{0.43} & 83.27 \std{0.43}  & 86.66 \std{0.48}\\
    \midrule
    \multicolumn{5}{l}{\textbf{\textit{Hyperparameter Tuning and Architecture Variation:}}} \\
    
    \multicolumn{1}{l|}{Dropout \textbf{(2)}}                     & 54.44 \std{0.62} & 55.41 \std{0.32} & 64.55 \std{0.45}  & 83.04 \std{0.34} & 83.65 \std{0.29} & 86.56 \std{0.44} \\
    \multicolumn{1}{l|}{Depth=1,2 \textbf{(2)}}  & 54.58 \std{0.52} & 55.22 \std{0.45} & 60.89 \std{0.41}  & 82.47 \std{0.44} & 83.03 \std{0.45} & 89.22 \std{0.41}\\
    \multicolumn{1}{l|}{Depth=1,2,6 \textbf{(3)}}             & 54.92 \std{0.47} & 55.56 \std{0.38} & 64.43 \std{0.47} & 82.47 \std{0.43} & 83.41 \std{0.52} & 91.95 \std{0.46} \\
    \multicolumn{1}{l|}{Filter=$A_{\text{rw}/\text{sym}}$,$A_{\text{attn}}$ \textbf{(2)}}        & 54.76 \std{0.53} & 55.39 \std{0.53} & 61.26 \std{0.50}  & 83.54 \std{0.42} & 84.12 \std{0.38}  & 89.37 \std{0.45} \\
    \multicolumn{1}{l|}{Filter=$A_{\text{rw}/\text{sym}}$,$A_{\text{attn}}$,$L_{\text{sym}}$ \textbf{(3)}}        & 55.44 \std{0.50} & 55.83 \std{0.40} & 68.59 \std{0.38}  & 84.07 \std{0.86} & 85.23 \std{0.45}  & 95.48 \std{0.27} \\
    \midrule
    \multicolumn{5}{l}{\textbf{\textit{Training Data Selection (Domain Assignment Functions):}}} \\
    
    \multicolumn{1}{l|}{Random \textbf{(2)}}                     & 52.26 \std{0.46} & 53.97 \std{0.48} & 67.28 \std{0.56}  & 81.06 \std{0.57} & 82.76 \std{0.61} & 89.87 \std{0.63} \\
    \rowcolor{gray!10}\multicolumn{1}{l|}{Homophily \textbf{(2)}}             & 55.45 \std{0.28} & 58.10 \std{0.34} & 73.29 \std{0.46} & 76.74 \std{1.67} & 83.33 \std{0.45} & 94.03 \std{0.26} \\
    \rowcolor{gray!10}\multicolumn{1}{l|}{Homophily \textbf{(3)}}        & 53.37 \std{0.64} & 57.40 \std{0.40} & 79.34 \std{0.33}  & 83.42 \std{0.34} & 85.48 \std{0.33}  & 97.61 \std{0.13} \\
    \rowcolor{gray!10}\multicolumn{1}{l|}{Homophily-FT \textbf{(3)}}        & 57.89 \std{0.51} & 58.86 \std{0.45} & 74.70 \std{0.59}  & 84.92 \std{0.39} &  86.11 \std{0.32} & 96.06 \std{0.18} \\
    \multicolumn{1}{l|}{Filtered-Homophily-FT \textbf{(3)}}        & 55.91 \std{0.63} & 56.44 \std{0.66} & 73.59 \std{0.57}  & 84.55 \std{0.47} & 85.37 \std{0.44}  & 95.73 \std{0.31} \\

    
    \bottomrule
    \end{tabular}
    }
    \label{tab:downstream-task}
\end{table*}
We assess the effect of various expert diversification techniques on downstream node classification performance. For each technique, we train a set of diverse experts and aggregate their predictions using both Global Ensemble (Ensemble) and Moscat Gating Ensemble (MoE) approaches. We also report the Upper-bound Accuracy, which reflects the performance of an oracle gating model. All experiments are conducted with GCN as the base expert architecture. The experiment results are shown in Table ~\ref{tab:downstream-task}.

\section{Future Directions}

\noindent \textbf{Learning Domain Assignments.}
While our study focuses on heuristic-based domain assignments, future work should explore \textit{learnable} or \textit{adaptive} methods for expert diversification. This could involve clustering node embeddings, optimizing domain assignments via bilevel objectives, or jointly training the assignment and gating functions to improve alignment with task-specific signals.

\noindent \textbf{Robust Gating Models.}
The gating model can serve as a calibrator of expert confidence. However, its effectiveness depends on proper training. Future work can explore domain-aware gating objectives, uncertainty penalties, or curriculum-inspired data sampling to ensure the gating model generalizes across all regions of the graph.
The gating model can serve as a calibrator of expert confidence. However, its effectiveness depends on proper training. Future work can explore domain-aware gating objectives, uncertainty penalties, or curriculum-inspired data sampling to ensure the gating model generalizes across all regions of the graph.


\noindent \textbf{Improving Datasets and Evaluation.}
Current node classification benchmarks often limit meaningful expert-based GNN evaluation due to issues like synthetic/small-scale graphs, label sensitivity (e.g., Arxiv-Year), class imbalance (e.g., Genius, Questions), and shallow features that favor language models. 
Future benchmarks should include richer features, balanced labels, and diverse structures. Evaluation should reflect real-world heterogeneity, domain shifts, and use standardized metrics for specialization and generalization.

\end{document}